\documentclass[10pt,twocolumn,letterpaper]{article}

\usepackage{cvpr}
\usepackage{times}
\usepackage{epsfig}
\usepackage{graphicx}
\usepackage{amsmath}
\usepackage{amssymb}
\usepackage{layout}
\usepackage[nolist]{acronym} 

\usepackage{subfigure,times,multirow}
\usepackage{array}
\usepackage[dvipsnames]{xcolor}
\usepackage[ruled]{algorithm2e}
\usepackage{multirow}
\usepackage{cite}
\usepackage{kotex}
\usepackage[normalem]{ulem} 
\usepackage[keeplastbox]{flushend}
\usepackage[verbose=silent]{microtype}

\def\etal{\emph{et al}.}

\usepackage{pifont}
\newif\ifdraft\drafttrue
\ifdraft

\else

\fi

\usepackage{hyperref}

\cvprfinalcopy 

\def\cvprPaperID{892} 

\ifcvprfinal\fi

\begin{document}

\begin{acronym}[TRACA] 
\acro{TRACA}{TRAcker based on Context-aware deep feature compression with multiple Auto-encoders}
\end{acronym}

\title{Context-aware Deep Feature Compression for High-speed Visual Tracking}
\vspace{-3cm}
\author{Jongwon Choi$^1$ \quad Hyung Jin Chang$^{2,3}$ \quad Tobias Fischer$^2$ \quad Sangdoo Yun$^{1,4}$ \\ Kyuewang Lee$^1$ \quad Jiyeoup Jeong$^1$ \quad Yiannis Demiris$^2$ \quad Jin Young Choi$^1$\\
{\small \hspace{1cm}$^1$ASRI, ECE., Seoul National University}
{\small \hspace{1cm}$^2$Personal Robotics Lab., EEE., Imperial College London}\\
{\small \hspace{-1.5cm}$^3$School of Computer Science, University of Birmingham}
{\small \hspace{1.5cm}$^4$Clova AI Research, NAVER Corp.}\\
{\tt\scriptsize jwchoi.pil@gmail.com, \{hj.chang,t.fischer,y.demiris\}@imperial.ac.uk, \{yunsd101,kyuewang,jy.jeong,jychoi\}@snu.ac.kr}
}

\maketitle
\thispagestyle{empty}

\AddToShipoutPicture*{%
     \AtTextUpperLeft{%
         \put(0,30){
           \begin{minipage}{\textwidth}
              \footnotesize
              Preprint version; final version available at \url{http://ieeexplore.ieee.org}\\
              IEEE Conference on Computer Vision and Pattern Recognition (CVPR) (2018)\\
              Published by: IEEE
           \end{minipage}}%
     }%
}

\begin{abstract}
We propose a new context-aware correlation filter based tracking framework to achieve both high computational speed and state-of-the-art performance among real-time trackers.
The major contribution to the high computational speed lies in the proposed deep feature compression that is achieved by a context-aware scheme utilizing multiple expert auto-encoders; a context in our framework refers to the coarse category of the tracking target according to appearance patterns. 
%
In the pre-training phase, one expert auto-encoder is trained per category.
In the tracking phase, the best expert auto-encoder is selected for a given target, and only this auto-encoder is used.
To achieve high tracking performance with the compressed feature map, we introduce extrinsic denoising processes and a new orthogonality loss term for pre-training and fine-tuning of the expert auto-encoders.
We validate the proposed context-aware framework through a number of experiments, where our method achieves a comparable performance to state-of-the-art trackers which cannot run in real-time, while running at a significantly fast speed of over 100 fps.
\end{abstract}
\section{Introduction}
The performance of visual trackers has vastly improved with the advances of deep learning research.
Recently, two different groups for deep learning based tracking have emerged.
The first group consists of online trackers which rely on continuous fine-tuning of the network to learn the changing appearance of the target~\cite{ref:FCNT,ref:MDNet,ref:SINT,ref:STCT,ref:yun}.
While these trackers result in high accuracy and robustness, their computational speed is insufficient to fulfil the real-time requirement of online tracking.
The second group is composed of correlation filter based trackers utilising raw deep convolutional features~\cite{ref:HDT,ref:CF2,ref:DeepSRDCF,ref:COT,ref:ECO}.
However, these features are designed to represent general objects contained in large datasets such as ImageNet~\cite{ref:ImageNet} and therefore are of high dimensionality. 
As the computational time for the correlation filters increases with the feature dimensionality, trackers within the second group do not satisfy the real-time requirement of online tracking either.

\begin{figure}[t]
    \includegraphics[width=1.0\linewidth]{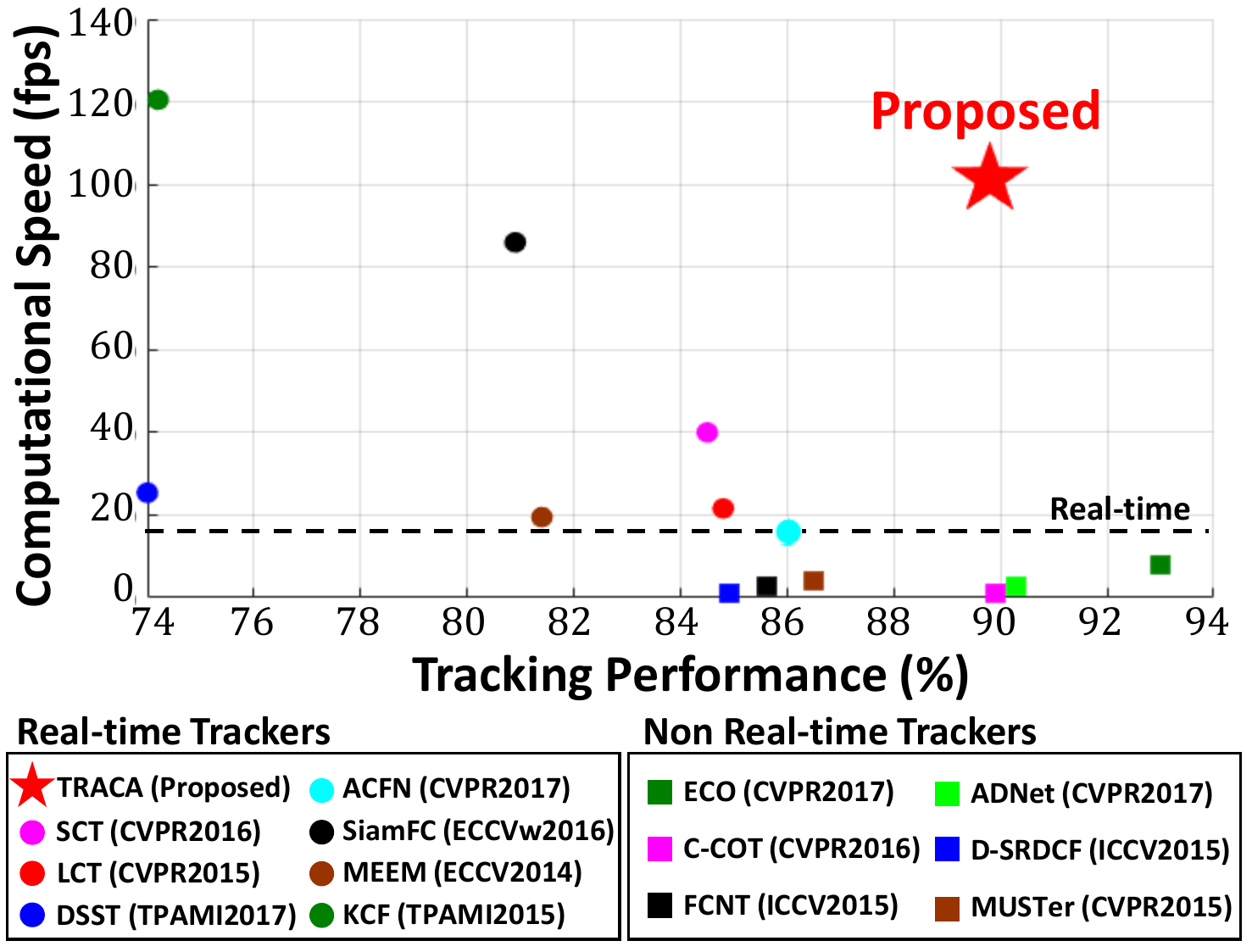}
    \caption{ \textbf{Comparison of computational efficiency.} This plot compares the performance and computational speed of the proposed tracker (TRACA) with previous state-of-the-art trackers using the CVPR2013 dataset~\cite{ref:Benchmark}. TRACA shows comparable performance with the best performing non real-time trackers, while running at a fast speed of over 100 fps.
    }
    \label{fig:introduction}
    \vspace{-0.4cm}
\end{figure}


In this work, we propose a correlation filter based tracker using context-aware compression of raw deep features, which reduces computational time, thus increasing speed.
This is motivated by the observation that a lower dimensional feature map can sufficiently represent the single target object which is in contrast to the classification and detection tasks using large datasets that cover numerous object categories.
Compression of high dimensional features into a low dimensional feature map is performed using autoencoders~\cite{ref:dAE, ref:ae1, ref:ae2, ref:ae3}.
More specifically, we employ multiple auto-encoders whereby each auto-encoder specialises in a specific category of objects; these are referred to as \textit{expert auto-encoders}.
We introduce an unsupervised approach to find the categories by clustering the training samples according to contextual information, and subsequently train one expert auto-encoder per cluster.
During visual tracking, an appropriate expert auto-encoder is selected by a context-aware network given a specific target. 
The compressed feature map is then obtained after fine-tuning the selected expert auto-encoder by a novel loss function considering the orthogonality of the correlation filters.
The compressed feature map contains reduced redundancy and sparsity, which increases accuracy and computational efficiency of the tracking framework.
%
To track the target, correlation filters are applied to the compressed feature map.
We validate the proposed framework through a number of self-comparisons and show that it outperforms other trackers using raw deep features while being notably faster at a speed of over 100 fps (see Fig.~\ref{fig:introduction}).

\begin{figure*}[t]
\centering
    \includegraphics[width=0.82\linewidth]{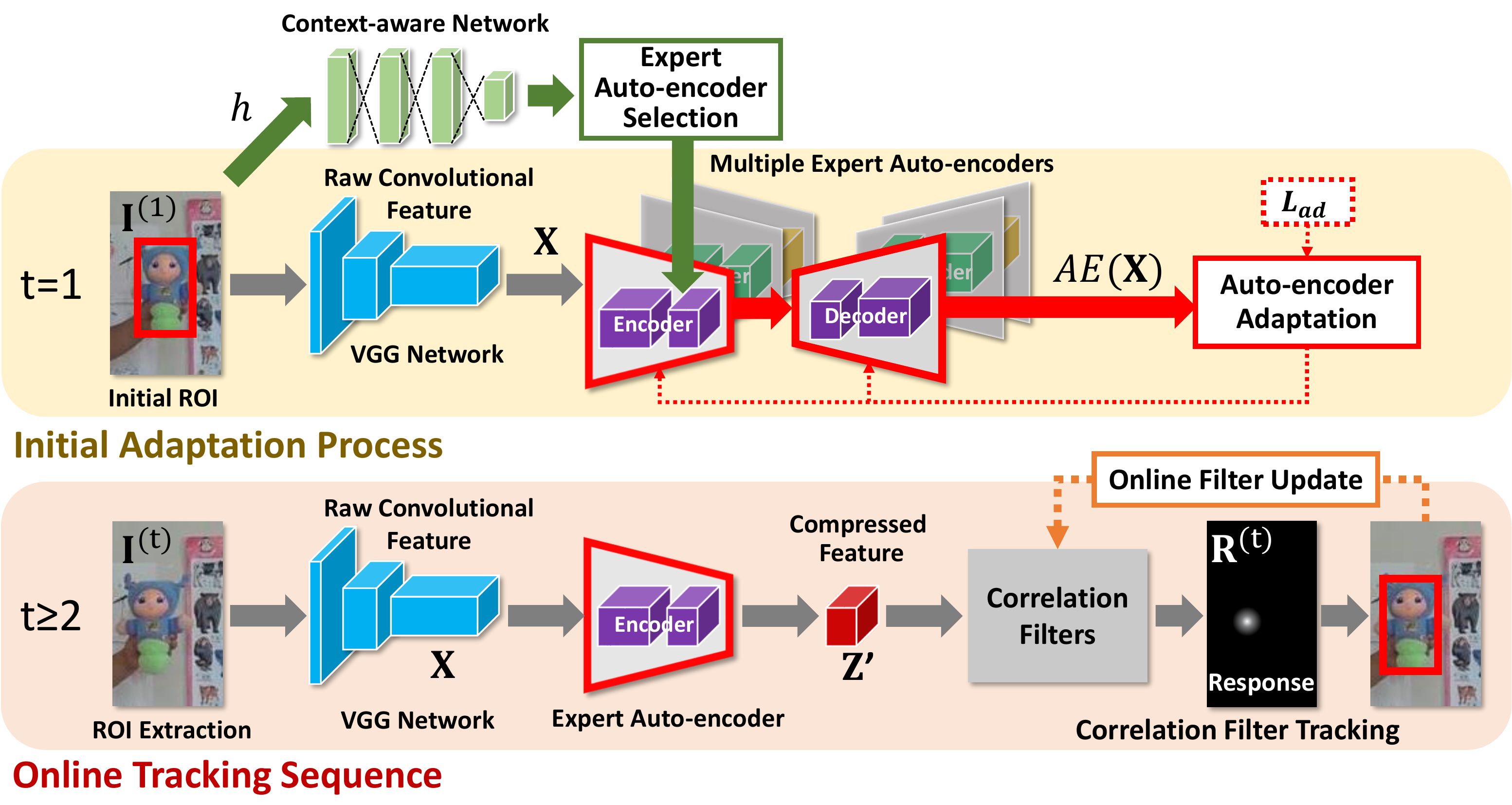}
    \vspace{-0.1cm}
    \caption{ {\bf{Proposed algorithm scheme.}} The expert auto-encoder is selected by the context-aware network and fine-tuned once by the ROI patch at the initial frame ($\mathbf{I}^{(1)}$). For the following frames, we first extract the ROI patch ($\mathbf{I}^{(t)}$) centred at the previous target position. Then, a raw deep convolutional feature ($\mathbf{X}$) is obtained through VGG-Net, and is compressed by the fine-tuned expert auto-encoder. The compressed feature ($\mathbf{Z}'$) is used as the feature map for the correlation filter, and the target's position is determined by the peak position of the filter response. After each frame, the correlation filter is updated online by the newly found target's compressed feature.}
    \label{fig:framework}
    \vspace{-0.4cm}
\end{figure*}

\section{Related Works}

\textbf{Online deep learning based trackers:}
Recent trackers based on online deep learning~\cite{ref:FCNT, ref:MDNet,ref:SINT,ref:STCT,ref:yun} have outperformed previous low-level feature-based trackers. 
Wang~\etal~\cite{ref:FCNT} proposed a framework simultaneously utilising shallow and deep convolutional features to consider detailed and contextual information of the target respectively.
Nam and Han~\cite{ref:MDNet} introduced a novel training method which
avoids overfitting by appending a classification layer to a convolutional neural network that is updated online. 
Tao~\etal~\cite{ref:SINT} utilised a Siamese network to estimate the similarities between the target's previous appearance and the current candidate patches.
Yun~\etal~\cite{ref:yun} suggested a new tracking method using an action decision network
 which can be trained by a reinforcement learning method with weakly labelled datasets.
However, trackers based on online deep learning require frequent fine-tuning of the networks, which is slow and prohibits real-time tracking.
David~\etal~\cite{ref:goturn} and Bertinetto~\etal~\cite{ref:SiamFC} proposed pre-trained networks to quickly track the target without online fine-tuning, but the performance of these trackers is lower than that of the state-of-the-art trackers.

\textbf{Correlation filter based trackers:}
The correlation filter based approach for visual tracking has become increasingly popular due to its rapid computation speed~\cite{ref:KCF, ref:MOSSE, ref:DSST, ref:LongCT, ref:MUSTer, ref:SCT, ref:ACFN}.
Henriques~\etal~\cite{ref:KCF} improved the tracking performance by extending the correlation filter to multi-channel inputs and kernel-based training.
Danelljan~\etal~\cite{ref:DSST} developed a new correlation filter that can detect scale changes of the target. Ma~\etal~\cite{ref:LongCT} and Hong~\etal~\cite{ref:MUSTer} integrated correlation filters with an additional long-term memory system.
Choi~\etal~\cite{ref:ACFN} proposed a tracker with an attentional mechanism exploiting previous target appearance and dynamics.

Correlation filter based trackers showed state-of-the-art performance when deep convolutional features were utilised~\cite{ref:HDT,ref:DeepSRDCF,ref:COT,ref:ECO}.
Danelljan~\etal~\cite{ref:DeepSRDCF} extended the regularised correlation filter~\cite{ref:SRDCF} to use deep convolutional features.
Danelljan~\etal~\cite{ref:COT} also proposed a novel correlation filter to find the target position in the continuous domain to incorporate features of various resolutions.
Ma~\etal~\cite{ref:HDT} estimated the target position by fusing the response maps obtained from convolutional features of various resolutions.
However, even though each correlation filter works fast, raw deep convolutional features have too many channels to be handled in real-time.
A first step towards decreasing the feature space was made by Danelljan~\etal~\cite{ref:ECO} by considering the linear combination of raw deep features, however the method still cannot run in real-time, and the deep feature redundancy was not fully suppressed.


\textbf{Multiple-context deep learning frameworks:}
Our proposed tracking framework benefits from the observation that the performance of deep networks can be improved using contextual information to train multiple specialised deep networks. Indeed, there are several works utilizing such a scheme.
Li~\etal~\cite{ref:context_cascadeface} proposed a cascaded framework detecting faces through multiple neural networks trained by samples divided according to the degree of their detection difficulty.
Vu~\etal~\cite{ref:context_head} integrated the head detection results from two neural networks, one specialising in local information and the other one in global information.
Neural networks specialising in local and global information have also been utilised in the saliency map estimation task~\cite{ref:context_saliency, ref:context_saliency2}.
In crowd density estimation, many works~\cite{ref:crowd1, ref:crowd2, ref:crowd3} have increased their performance by using multiple deep networks with different receptive fields to cover various scales of crowds.

\section{Methodology}
The proposed \ac{TRACA} consists of multiple expert auto-encoders, a context-aware network, and correlation filters as shown in Fig.~\ref{fig:framework}. 
The expert auto-encoders robustly compress raw deep convolutional features from VGG-Net~\cite{ref:vggm}.
Each of them is trained according to a different context, and thus performs context-dependent compression (see Sec.~\ref{sec:expertae}).
We propose a context-aware network to select the expert auto-encoder best suited for the specific tracking target, and only this auto-encoder is running during online tracking (see Sec.~\ref{sec:contextnet}).
After initially adapting the selected expert auto-encoder for the tracking target, its compressed feature map is utilised as an input of correlation filters which track the target online. 
We introduce the general concept of correlation filters in Sec.~\ref{sec:correlationfilter} and then detail the tracking processes including the initial adaptation and the online tracking in Sec.~\ref{Sec:tracking}.

\subsection{Expert Auto-encoders} \label{sec:expertae}

\textbf{Architecture:}
Auto-encoders have shown to be suitable for unsupervised feature learning~\cite{ref:ae_hinton2006,ref:ae_hinton2006_2,ref:dAE}.
They offer a way to learn a compact representation of the input while retaining the most important information to recover the input given the compact representation.
In this paper, we propose to use a set of $N_e$ expert auto-encoders of the same structure, each covering a different context. 
The inputs to be compressed are raw deep convolutional feature maps obtained from one of the convolution layers in VGG-Net~\cite{ref:vggm}.

To achieve a high compression ratio, we stack $N_l$ encoding layers which are followed by $N_l$ decoding layers in the auto-encoder.
The $l$-th encoding layer $f_l$ is a convolutional layer working as $f_l: \mathbb{R}^{w\times h\times c_l} \rightarrow \mathbb{R}^{w\times h \times c_{l+1}}$, thus reducing the channel dimension $c_l$ of the input to latent channel dimension $c_{l+1}$ while preserving the resolution of the feature map.
The output of $f_l$ is provided as input to $f_{l+1}$ such that the channel dimension $c$ decreases as the feature maps pass through the encoding layers.
More specifically, in our proposed framework one encoding layer reduces the channel dimension in half, \ie $c_{l+1}=c_l/2$ for $l\in\{1,\cdots,N_l\}$.
By denoting the ($N_l-k+1)$-th decoding layer by $g_k$ in the adverse way of $f_l$, $~g_k: \mathbb{R}^{w\times h\times c_{k+1}} \rightarrow \mathbb{R}^{w\times h \times c_k}$ expands the input channel dimension $c_{k+1}$ into $c_k$ to restore the original dimension $c_1$ of $\mathbf{X}$ at the last layer of the decoder, where $k\in\{1,\cdots,N_l\}$. 
Then, the auto-encoder $AE$ can be expressed as $AE(\mathbf{X}) \equiv g_1(\cdots (g_{N_l}(f_{N_l}(\cdots(f_1(\mathbf{X})))))\in\mathbb{R}^{w\times h\times c_1}$ for a raw convolutional feature map
$\mathbf{X}\in\mathbb{R}^{w\times h\times c_1}$, and the compressed feature map in the auto-encoder is defined as $\mathbf{Z}\equiv f_{N_l}(\cdots(f_1(\mathbf{X})))\in\mathbb{R}^{w\times h\times c_{N_l+1}}$.
All convolution layers are followed by the ReLU activation function, and the size of their convolution filters is set to $3\times 3$.

\textbf{Pre-training:}
The pre-training phase for the expert auto-encoders is split into three parts, each serving a distinct purpose. 
First, we train the base auto-encoder $AE^o$ using all training samples to find context-independent initial compressed feature maps. 
Then, we perform contextual clustering on the initial compressed feature maps of $AE^o$ to find $N_e$ context-dependent clusters. Finally, these clusters are used to train the expert auto-encoders initialised by the base auto-encoder with one of the sample clusters. 

The purpose of the base auto-encoder is twofold: Using the context-independent compressed feature maps to cluster the training samples and finding good initial weight parameters from which the expert auto-encoders can be fine-tuned.
The base auto-encoder is trained by 
raw convolutional feature maps $\{ \mathbf{X}_j \}_{j=1}^m$ with a batch size $m$.
The $\mathbf{X}_j$ is obtained as the output from a convolutional layer involved in VGG-Net~\cite{ref:vggm} fed by randomly selected training images $\mathbf{I}_j$ from a large image database such as ImageNet~\cite{ref:ImageNet}.

\begin{figure}[t]
    \subfigure[Channel corrupting process]{\includegraphics[width=4.0cm]{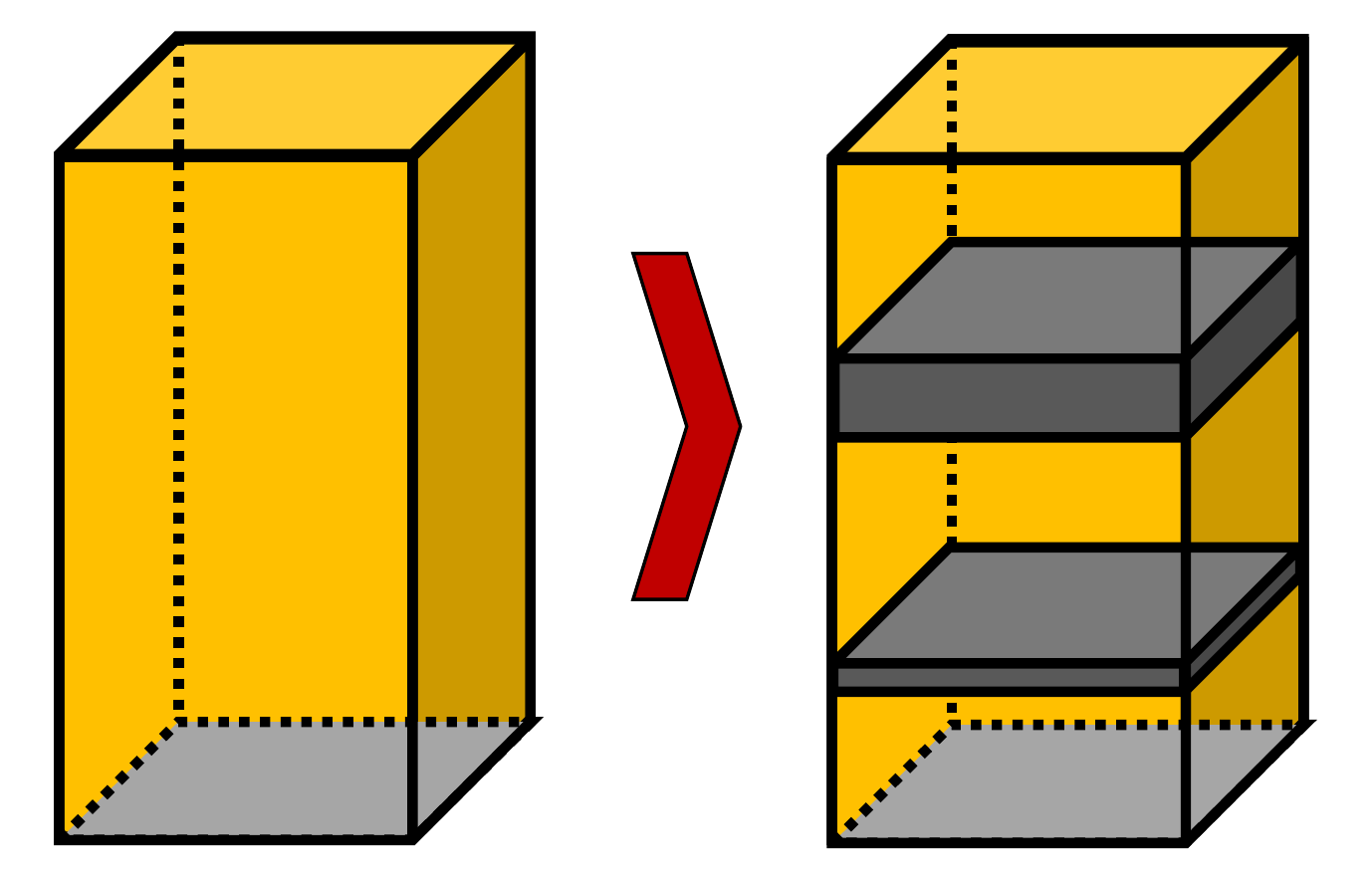}}
    \hfill
    \subfigure[Feature vector exchange process]{\includegraphics[width=4.0cm]{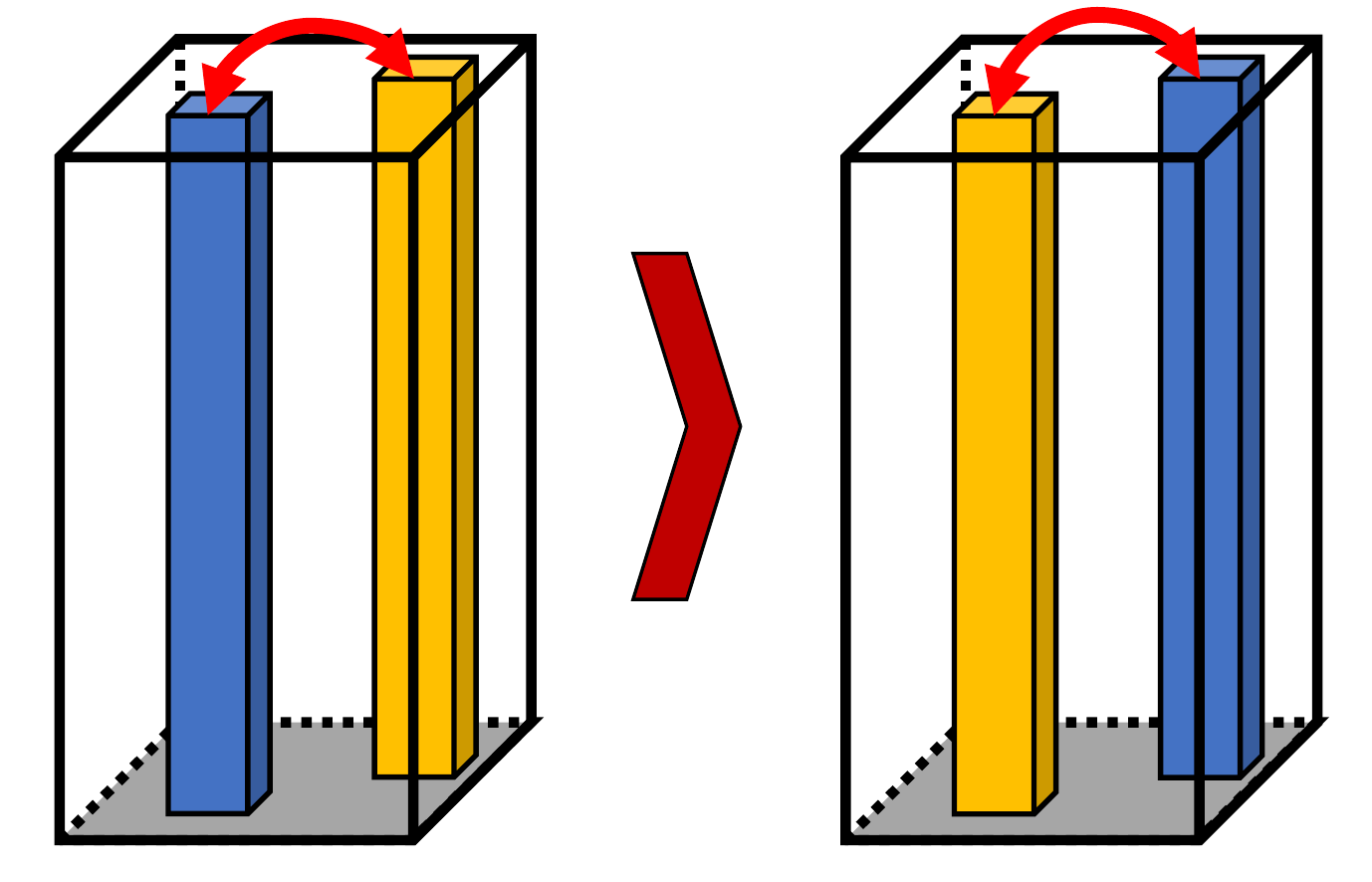}}
    \caption{\textbf{Extrinsic denoising criteria.} To increase robustness of the compressed feature map in the pre-training, two extrinsic denoising criteria are applied to the raw deep feature map which is the input of the auto-encoder. (a) In the channel corrupting process, some randomly selected channels are set to zero. (b) In the exchange process, randomly chosen feature vectors are interchanged.}
    \label{fig:denoising}
    \vspace{-0.4cm}
\end{figure}

To make the base auto-encoder more robust to appearance changes and occlusions, we use two denoising criteria which help to capture distinct structures in the input distribution (illustrated in Fig.~\ref{fig:denoising}). 
The first denoising criterion is a {\it channel corrupting process} where a fixed number of feature channels is randomly chosen and the values for these channels is set to $0$ (while the other channels remain unchanged), which is similar to the destruction process of denoising auto-encoders~\cite{ref:dAE}.
Thus all information for these channels is removed and the auto-encoder is trained to recover this information. 
The second criterion is an {\it exchange process}, where some spatial feature vectors of the convolutional feature are randomly interchanged.
Since the receptive fields of the feature vectors cover different regions within an image, exchanging the feature vectors is similar to interchanging regions within the input image. Thus, interchanging feature vectors that cover the background region and target region respectively leads to a similar effect as the background occluding the target. Therefore, the auto-encoders are trained to be robust against occlusions.
We denote $\{ \check{\mathbf{X}}_j \}_{j=1}^m$ as the mini-batch after performing the two denoising processes.
Then, the base auto-encoder $AE^o$ can be trained by minimising the distance between the input feature map $\mathbf{X}_j$ and its output $AE^o(\check{\mathbf{X}}_j)$ with the noisy sample $\check{\mathbf{X}}_j$. 

However, when we only consider the distance between the input and the final output of the base auto-encoder, we frequently observed an overfitting problem and unstable training convergence. To solve these problems, we design a novel loss based on a multi-stage distance which consists of the distances between the input and the outputs obtained by the partial auto-encoders. The partial auto-encoders $\{AE_i(\mathbf{X})\}_{i=1}^{N_l}$ contain only a portion of the encoding and decoding layers of their original auto-encoder $AE(\mathbf{X})$, while the input and output sizes match that of the original auto-encoder, \ie\ $AE_1(\mathbf{X})=g_1(f_1(\mathbf{X}))$, $AE_2(\mathbf{X})=g_1(g_2(f_2(f_1(\mathbf{X}))))$, $\cdots$ when $AE(\mathbf{X}) = g_1(\cdots (g_{N_l}(f_{N_l}(\cdots(f_1(\mathbf{X}))))))$. Thus, the loss based on the multi-stage distance can be described as:
\small
\begin{equation}
L_{ae} = \frac{1}{m} \sum_{j=1}^m \sum_{i=1}^{N_l} \| \mathbf{X}_j - AE^o_i(\check{\mathbf{X}}_j) \|_2^2,
\end{equation}
\normalsize
where $AE^o_i(\mathbf{X})$ is the $i$-th partial auto-encoder of $AE^o(\mathbf{X})$, and recall that $m$ denotes the mini batch size.


Then, we cluster the training samples $\{ \mathbf{I}_j \}_{j=1}^N$ according to their respective feature maps compressed by the base auto-encoder, where $N$ denotes the total number of training samples.
To avoid overfitting of the expert auto-encoders due to a too small cluster size, we introduce a two-step clustering algorithm which avoids small clusters.

In the first step, we find $2N_e$ samples which are chosen randomly from the feature maps compressed by the base auto-encoder (note that this is twice the amount of desired clusters). 
We repeat the random selection 1000 times and find the samples which have the largest Euclidean distance amongst them as initial centroids.
Then, all training samples are clustered by $k$-means clustering with $k = 2N_e$ using the compressed feature maps. 
In the second step, among the resulting $2N_e$ centroids, we remove the $N_e$ centroids of the clusters with the smallest number of included samples. 
Then, $N_e$ centroids remain, and we cluster the training samples again using these centroids, which results in $N_e$ clusters including enough samples to avoid the overfitting problem.
We denote the cluster index for $\mathbf{I}_j$ as $d_j\in\{1,...,N_e\}$.

The $d$-th expert auto-encoder $AE^d$ is then found by fine-tuning the base auto-encoder using the training samples with contextual cluster index $d$.
The training process (including the denoising criteria) differs from the base auto-encoder only in the training samples.

\subsection{Context-aware Network} \label{sec:contextnet}

\textbf{Architecture:}
The context-aware network selects the expert auto-encoder which is most contextually suitable for a given tracking target.
We adopt a pre-trained VGG-M model~\cite{ref:vggm} for the context-aware network since it contains a large amount of semantic information from pre-training on ImageNet~\cite{ref:ImageNet}. 
Given a $224\times 224$ RGB input image, our context-aware network consists of three convolutional layers \{\textit{conv1}, \textit{conv2}, \textit{conv3}\} followed by three fully connected layers \{\textit{fc4}, \textit{fc5}, \textit{fc6}\}, whereby \{\textit{conv1}, \textit{conv2}, \textit{conv3}, \textit{fc4}\} are identical to the corresponding layers in VGG-M. 
The weight parameters of \textit{fc5} and \textit{fc6} are initialised randomly with zero-mean Gaussian distribution. 
\textit{fc5} is followed by a ReLU function and contains 1024 output nodes. 
Finally \textit{fc6} has $N_e$ output nodes and is combined with a softmax layer to estimate the probability for each of the expert auto-encoders to be suited for the tracking target.

\textbf{Pre-training:}
The context-aware network takes a training sample $\mathbf{I}_j$ as input and outputs the estimated probabilities of that sample belonging to cluster index $d_j$.
It is being trained by a batch $\{ \mathbf{I}_j, d_j \}_{j=1}^{m'}$ of image/cluster-index pairs where $m'$ is the mini-batch size for the context-aware network.
We fix the weights of \{\textit{conv1}, \textit{conv2}, \textit{conv3}, \textit{fc4}\}, and train the weights for \{\textit{fc5}, \textit{fc6}\} by minimising the multi-class loss function $L_{pr}$ using stochastic gradient descent, where
\small
\vspace{-2mm}
\begin{equation}
L_{pr} = \frac{1}{m'} \sum_{j=1}^{m'} H(d_j, h(\mathbf{I}_j)),
\vspace{-2mm}
\end{equation}
\normalsize
$H$ denotes the cross-entropy loss, and $h(\mathbf{I}_j)$ is the predicted cluster index of $\mathbf{I}_j$ by the context-aware network $h$.

\subsection{Correlation Filter} \label{sec:correlationfilter}
Before detailing the tracking process of TRACA, we briefly introduce the functionality of conventional correlation filters using a single-channel feature map.
Based on the property of the circulant matrix in the Fourier domain~\cite{ref:cirMatrix}, correlation filters can be trained quickly which leads to high-performing trackers under low computational load~\cite{ref:KCF}. 
Given the vectorised single-channel training feature map $\mathbf{z}\in\mathbb{R}^{wh\times 1}$ 
and the vectorised target response map $\mathbf{y}$ obtained from a 2-D Gaussian window with size $w\times h$ and variance $\sigma_y^2$ as in \cite{ref:KCF}, the vectorised correlation filter $\mathbf{w}$ can be estimated by:
\begin{equation} \label{eq:cf}
\mathbf{w} = \mathcal{F}^{-1} \left( \frac{\widehat{\mathbf{z}} \odot \widehat{\mathbf{y}}}{\widehat{\mathbf{z}}\odot \widehat{\mathbf{z}}^* + \lambda} \right),
\end{equation}
where $\hat{\mathbf{y}}$ and $\hat{\mathbf{z}}$ represent the Fourier-transformed vector of $\mathbf{y}$ and $\widehat{\mathbf{z}}$ respectively, $\widehat{\mathbf{z}}^*$ is the conjugated vector of $\mathbf{z}$, $\odot$ denotes an element-wise multiplication, $\mathcal{F}^{-1}$ stands for an inverse Fourier transform function, and $\lambda$ is a predefined regularisation factor.

For the vectorised single-channel test feature map $\mathbf{z}'\in\mathbb{R}^{wh\times 1}$, the vectorised response map $\mathbf{r}$ can be obtained by:
\vspace{-2mm}
\begin{equation}\label{eq:cfrs}
\mathbf{r} = \mathcal{F}^{-1} \left( \widehat{\mathbf{w}} \odot \widehat{\mathbf{z}}^{'*}\right).
\vspace{-2mm}
\end{equation}
Then, after re-building a 2-D response map $\mathbf{R}\in\mathbb{R}^{w\times h}$ from $\mathbf{r}$, the target position is found from the maximum peak position of $\mathbf{R}$.

\subsection{Tracking Process}
\label{Sec:tracking}
To track a target in a scene, we rely on a correlation filter based algorithm using the compressed feature map of the expert auto-encoders as selected by the context-aware network. 
We describe the initial adaptation of the selected expert auto-encoder in Sec.~\ref{sec:initialAdap} followed by a presentation of the correlation filter based tracking algorithm in Sec.~\ref{sec:onlintTrack}.

\subsubsection{Initial Adaptation Process} \label{sec:initialAdap}
\hspace{3mm}The initial adaptation process contains the following parts. We first extract a region of interest (ROI) including the target from the initial frame, and the expert auto-encoder suitable for the target is selected by the context-aware network.
Then, the selected expert auto-encoder is fine-tuned using the raw convolutional feature maps of the training samples augmented from the ROI.
When we obtain the compressed feature map from the fine-tuned expert auto-encoder, some of its channels represent background objects rather than the target.
Thus, we introduce an algorithm to find and remove the channels which respond to the background objects.

\textbf{Region of interest extraction: }
The ROI is centred around the target's initial bounding box, and is 2.5 times bigger than the target's size to cover the area nearby. 
We then resize the ROI of width $W$ and height $H$ to $224\times 224$ in order to match the expected input size of the VGG-Net.
This results in the resized ROI $\mathbf{I}^{(1)}\in\mathbb{R}^{224\times 224\times 3}$ for the $rgb$ domain. 
For grey-scale images, the grey value is replicated three times to obtain $\mathbf{I}^{(1)}$.
The best expert auto-encoder for the tracking scene is selected according to the contextual information of the initial target using the context-aware network $h$, and we can denote this auto-encoder as $AE^{h(\mathbf{I}^{(1)})}$.

\textbf{Initial sample augmentation: }
Even though we use two denoising criteria as described earlier, we found that the compressed feature maps of the expert auto-encoders show a deficiency for targets which become blurry or are flipped. 
Thus, we augment $\mathbf{I}^{(1)}$ in several ways before fine-tuning the selected auto-encoder. 
To tackle the blurriness problem, four augmented images are obtained by filtering $\mathbf{I}^{(1)}$ with Gaussian filters with variances $\{0.5, 1.0, 1.5, 2.0\}$. Two more augmented images are obtained by flipping $\mathbf{I}^{(1)}$ around the vertical and horizontal axes respectively.
Then, the raw convolutional feature maps extracted from the augmented samples can be represented by $\{\mathbf{X}^{(1)}_j\}_{j=1}^7$.

\textbf{Fine-tuning: }
The fine-tuning of the selected auto-encoder differs from the pre-training process for the expert auto-encoders.
As there is a lack of training samples, the optimisation rarely converges when applying the denoising criteria.
Instead, we employ a correlation filter orthogonality loss $L_{ad}$ which considers the orthogonality of the correlation filters estimated from the compressed feature map of the expert auto-encoder, where $L_{ad}$ is defined as:
\vspace{-2mm}
\begin{equation}
\small
\hspace{-3mm} L_{ad} = \hspace{-1mm} \sum_{j=1}^7 \sum_{i=1}^{N_l} \hspace{-1mm} \left\{ \hspace{-1mm} \big\| \mathbf{X}^{(1)}_j \hspace{-1mm}-\hspace{-1mm} AE_i(\mathbf{X}^{(1)}_j) \big\|_2^2 + \lambda_{\Theta} \hspace{-2mm} \sum_{k,l=1}^{c^{i+1}} \hspace{-1mm} \Theta(\mathbf{w}_{jik}, \mathbf{w}_{jil}) \hspace{-1mm}\right\},
\end{equation}
\normalsize
where $\Theta(\mathbf{u}, \mathbf{v}) = (\mathbf{u}\cdot\mathbf{v})^2 / (\|\mathbf{u}\|_2^2 \|\mathbf{v}\|_2^2) $ and $\mathbf{w}_{jik}$ defines a vectorised correlation filter estimated by Eq.(\ref{eq:cf}) using the vectorised $k$-th channel of the compressed feature map $f_i(\cdots(f_1(\mathbf{X}^{(1)}_j)))$ from the selected expert auto-encoder. 
The correlation filter orthogonality loss allows increasing the interaction among the correlation filters as estimated from the different channels of the compressed feature maps. 
The fine-tuning is performed by minimising $L_{ad}$ using stochastic gradient descent.
The differentiation of $L_{ad}$ is described in Appendix A of the supplementary material.

\textbf{Background channel removal: } 
The compressed feature map $\mathbf{Z}^{\forall}$ can be obtained from the fine-tuned expert auto-encoder.
Then, we remove the channels within $\mathbf{Z}^{\forall}$ which have large responses outside of the target bounding box.
Those channels are found by estimating the channel-wise ratio of foreground and background feature responses.
First, we estimate the channel-wise ratio of the feature responses for channel $k$ as 
\begin{equation}
ratio^k = \|vec(\mathbf{Z}^{k, \forall}_{bb})\|_1 / \|vec(\mathbf{Z}^{k, \forall})\|_1,
\end{equation}
where $\mathbf{Z}^{k, \forall}$ is the $k$-th channel feature map of $\mathbf{Z}^{\forall}$ and $\mathbf{Z}^{k, \forall}_{bb}$ is obtained from $\mathbf{Z}^{k, \forall}$ by setting the values out of the target bounding box to $0$ while the other values are untouched. 
Then, after sorting all channels according to $ratio^k$ in descending order, only the first $N_c$ channels of the compressed feature map are utilised as input to the correlation filters. We denote the resulting feature map as $\mathbf{Z}\in\mathbb{R}^{S\times S\times N_c}$, where $S$ is the feature size.

\subsubsection{Online Tracking Sequence} \label{sec:onlintTrack}
\hspace{3mm}\textbf{Correlation filter estimation \& update: }
We first obtain the resized ROI for the current frame $t$ using the same method as in the initial adaptation, \ie\ the resized ROI is centred at the target's centre and its size is 2.5 times the target's size and resized to $224\times 224$.
After feeding the resized ROI to the VGG-Net, we obtain the compressed feature map $\mathbf{Z}^{(t)}\in\mathbb{R}^{S\times S\times N_c}$ by inserting the raw deep convolutional feature map of the VGG-Net into the adapted expert auto-encoder.

Then, using Eq.(\ref{eq:cf}), we estimate independent correlation filters $\mathbf{w}^{k,(t)}$ for each feature map $\mathbf{Z}^{k,(t)}$, where $\mathbf{Z}^{k,(t)}$ denotes the $k$-th channel of $\mathbf{Z}^{(t)}$.
Following~\cite{ref:KCF}, we suppress background regions by multiplying each $\mathbf{Z}^{k,(t)}$ with cosine windows of the same size.
For the first frame, we can estimate the correlation filters $\bar{\mathbf{w}}^{k,(1)}$ with Eq.(\ref{eq:cf}) given by $\mathbf{Z}^{k,(1)}$.
For the following frames ($t>1$), the correlation filters are updated as follows:
\begin{equation}
\bar{\mathbf{w}}^{k,(t)} = (1-\gamma)\bar{\mathbf{w}}^{k,(t-1)} + \gamma\mathbf{w}^{k,(t)},
\label{eq:update}
\end{equation}
where $\gamma$ is an interpolation factor.

\textbf{Tracking: }
After estimating the correlation filter, we need to find the position $[x^t, y^t]$ of the target in frame $t$. 
As we assume that $[x^t, y^t]$ is close to the target position in the previous frame $\left([x^{t-1}, y^{t-1}]\right)$, we extract the tracking ROI from the same position as the ROI for the correlation filter estimation of the previous frame.
Then, we can obtain the compressed feature map $\mathbf{Z}^{'(t)}$ for tracking using the adapted expert auto-encoder. 
Inserting $\mathbf{Z}^{'(t)}$ and $\bar{\mathbf{w}}^{k,(t-1)}$ to Eq.(\ref{eq:cfrs}) then provides the channel-wise response map $\mathbf{R}^{k,(t)}$ 
(we apply the multiplication of cosine windows in the same manner as for the correlation filter estimation).

We then need to combine $\mathbf{R}^{k,(t)}$ to the integrated response map $\mathbf{R}^{(t)}$. 
We use a weighted averaging scheme, where we use the validation score $s^k$ as weight factor with
\begin{equation}
s^k = \exp\left(-\lambda_s\| \mathbf{R}^{k,(t)} - \mathbf{R}_o^{k,(t)} \|_2^2\right),
\end{equation}
and $\mathbf{R}_o^{k,(t)}=\mathcal{G}(\mathbf{p}^{k,(t)},\sigma^2)_{S\times S}$ is a 2-D Gaussian window with size ${\small S\hspace{-1mm}\times\hspace{-1mm} S}$ and variance ${\small \sigma_y^2}$ centred at the peak point $\mathbf{p}^{k,(t)}$ of $\mathbf{R}^{k,(t)}$. 
Then, the integrated response map is calculated as:
\vspace{-2mm}
\begin{equation}
\mathbf{R}^{(t)} = \sum_{k=1}^{N_c} s^k \mathbf{R}^{k,(t)}.
\vspace{-2mm}
\end{equation}
Following~\cite{ref:ACFN}, we find the sub-pixel target position $\mathbf{p}^{(t)}$ by interpolating the response values near the peak position.
Finally, the target position $[x^t, y^t]$ is found as:
\begin{equation}
[x^t, y^t] = [x^{t-1}, y^{t-1}] + round( [W, H]\odot\mathbf{p}^{(t)}/S ).
\end{equation}

\textbf{Scale changes: }
To handle scale changes of the target, we extract two additional ROI patches scaled from the previous ROI patch size with scaling factors $1.015$ and $1.015^{-1}$ respectively in the tracking sequence. 
The new target scale is chosen as the scale where the respective maximum value of the response map (from the scaled ROI) is the largest.

\textbf{Full occlusion handling: }
To handle full occlusions, a re-detection algorithm is adopted. 
The overall idea is to introduce a so-called re-detection correlation filter which is not being updated and applied to the position of the target at the time where an occlusion has been detected. 
A full occlusion is assumed when a sudden drop of the maximum response value $R_{max}^{(t)}\equiv\max(\mathbf{R}^{(t)})$ is detected as described by $R_{max}^{(t)} < \lambda_{re} \bar{R}_{max}^{(t-1)}$ with $\bar{R}_{max}^{(t)}=(1-\gamma)\bar{R}_{max}^{(t-1)} + \gamma R_{max}^{(t)}$ and $\bar{R}_{max}^0 = R_{max}^1$ (note that this is the same $\gamma$ as in Eq.(\ref{eq:update})). 
If that condition is fulfilled, the correlation filter at time $(t-1)$ is used as re-detection correlation filter. 
During the next $N_{re}$ frames, the target position as determined by the re-detection correlation filter is being used if the maximum value of the re-detection filter's response map is larger than the maximum value of the response map obtained by the normal correlation filter. 
Furthermore, $\bar{\mathbf{w}}^{k,(t)}$ are replaced by the ones of the re-detection correlation filter, and the target scale is reset to the scale when the occlusion was detected.

\section{Experimental Result}

\subsection{Implementation}
The feature response after the second convolution layer ($conv2$) of VGG-M~\cite{ref:vggm} was given to the auto-encoders as raw convolutional feature input.
The number of expert auto-encoders was set to $N_e=10$, and their depth to $N_l=2$.
The mini-batch size for all auto-encoders was set to $m=10$.
The learning rate for the base auto-encoder was set to $10^{-10}$, and for expert auto-encoders to $10^{-9}$.
The base auto-encoder was trained for $10$ epochs, and the expert auto-encoders were fine-tuned for $30$ epochs.
The proportions for the two extrinsic denoising processes were set to $10\%$ respectively.
For training the context-aware network, the mini-batch size and the learning rate were set to $m'=100$ and $0.01$, respectively.
The weight for the orthogonality loss term was set to $\lambda_\Theta=10^3$, and the reduced channel dimension after the background channel removal was $N_c=25$.
The parameters for the correlation filter based tracker were set to $\sigma_g=0.05$, $\lambda=1.0$, and $\gamma=0.025$, and $\lambda_s=50$.
The parameters for full occlusion handling, $\lambda_{re}$ and $N_{re}$, were experimentally determined to $0.7$ and $50$ using scenes with occlusions.

We used MATLAB and MatConvNet~\cite{ref:matconvnet} to implement the proposed framework. 
The computational environment had an Intel i7-2700K CPU @ 3.50GHz, 16GB RAM, and an NVIDIA GTX1080 GPU. 
The computational speed was 101.3 FPS in the CVPR2013 dataset~\cite{ref:Benchmark}.
We release the source code along with the attached experimental results\footnote{\url{https://sites.google.com/site/jwchoivision/}}.

\begin{figure*}[t]
\centering
\subfigure{\includegraphics[width=5.7cm]{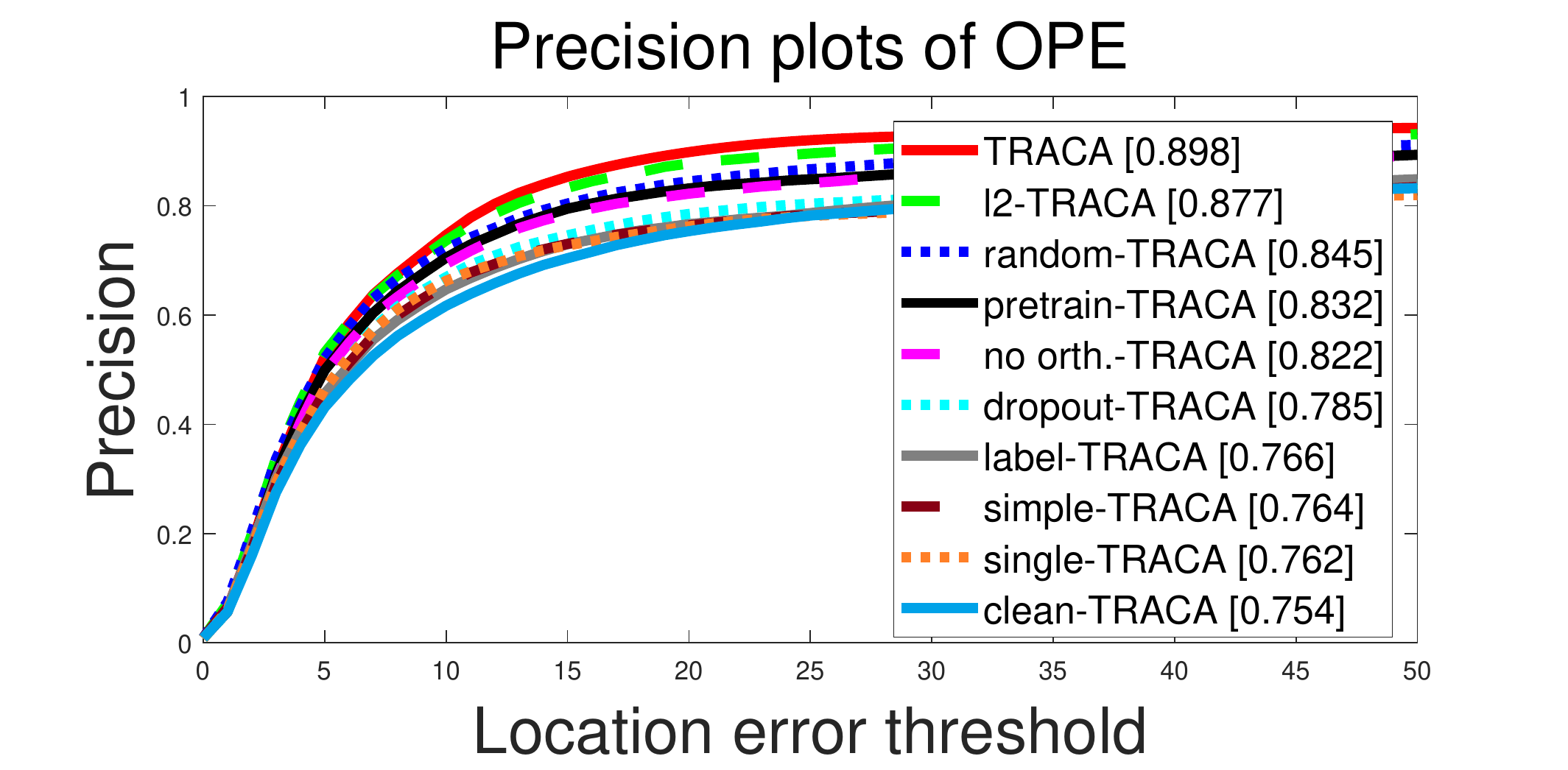}}
\hfill
\subfigure{\includegraphics[width=5.7cm]{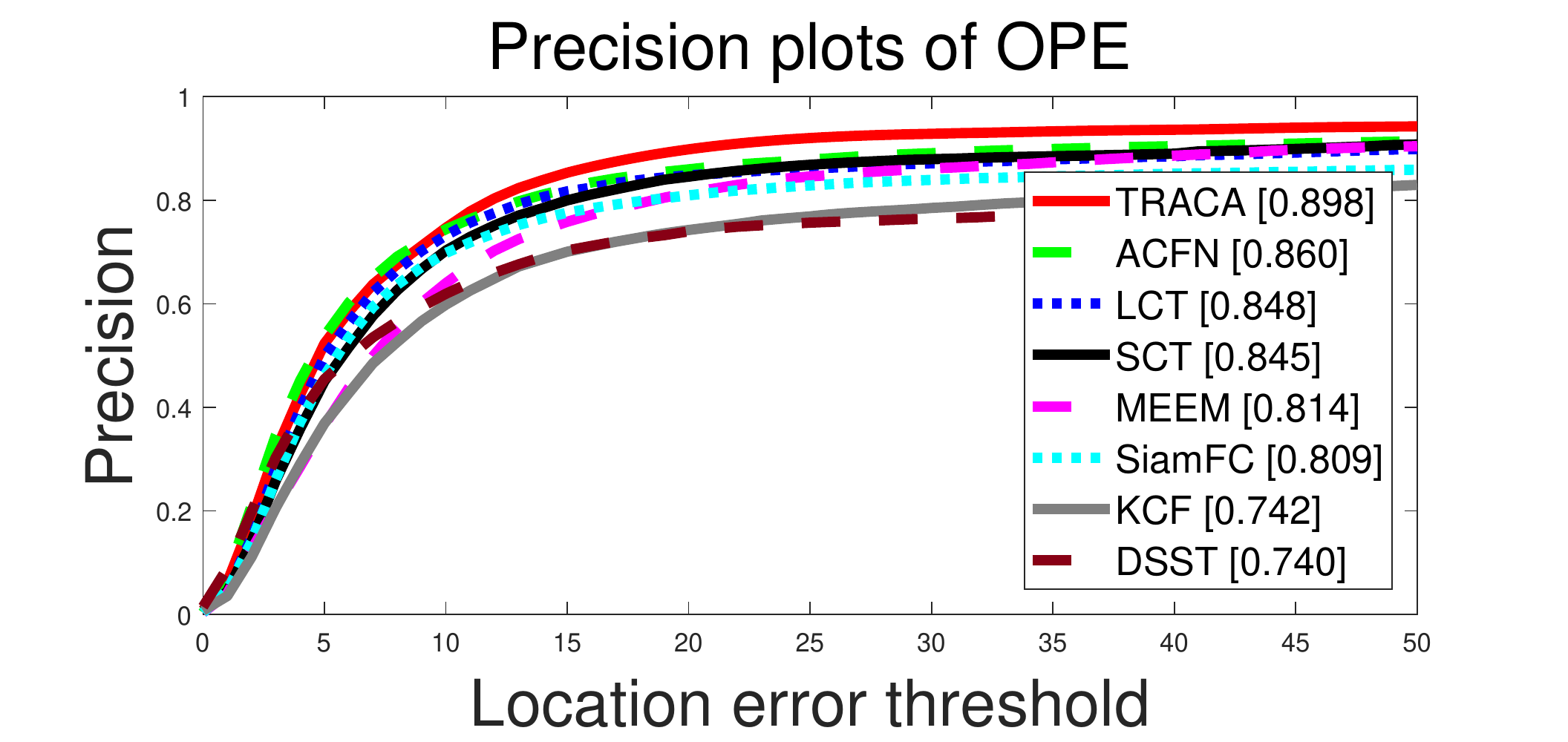}}
\hfill
\subfigure{\includegraphics[width=5.7cm]{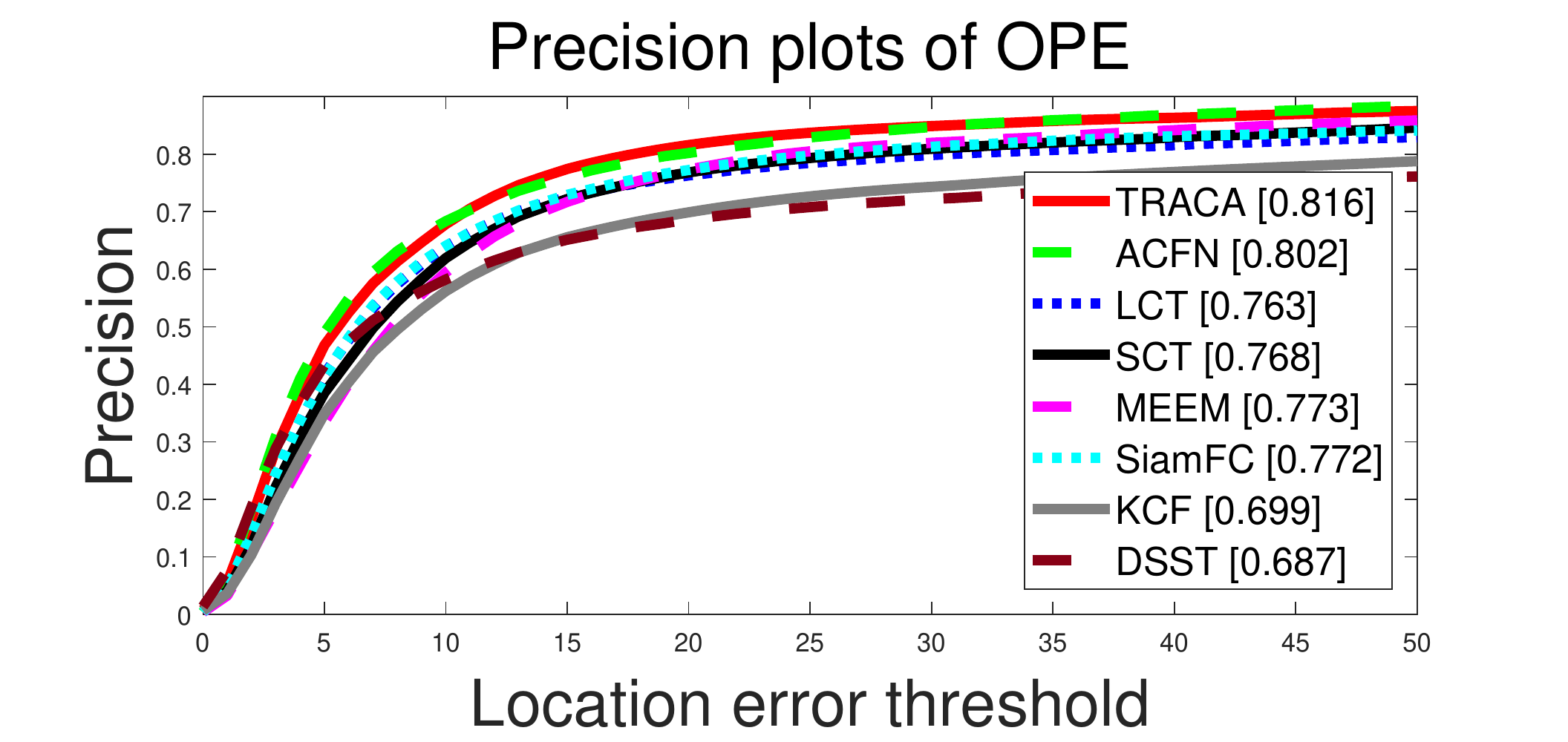}}
\addtocounter{subfigure}{-3}
\vspace{-0.6cm}
\end{figure*}
\begin{figure*}[t]
\centering
\subfigure[Self-comparison on CVPR2013]{\includegraphics[width=5.7cm]{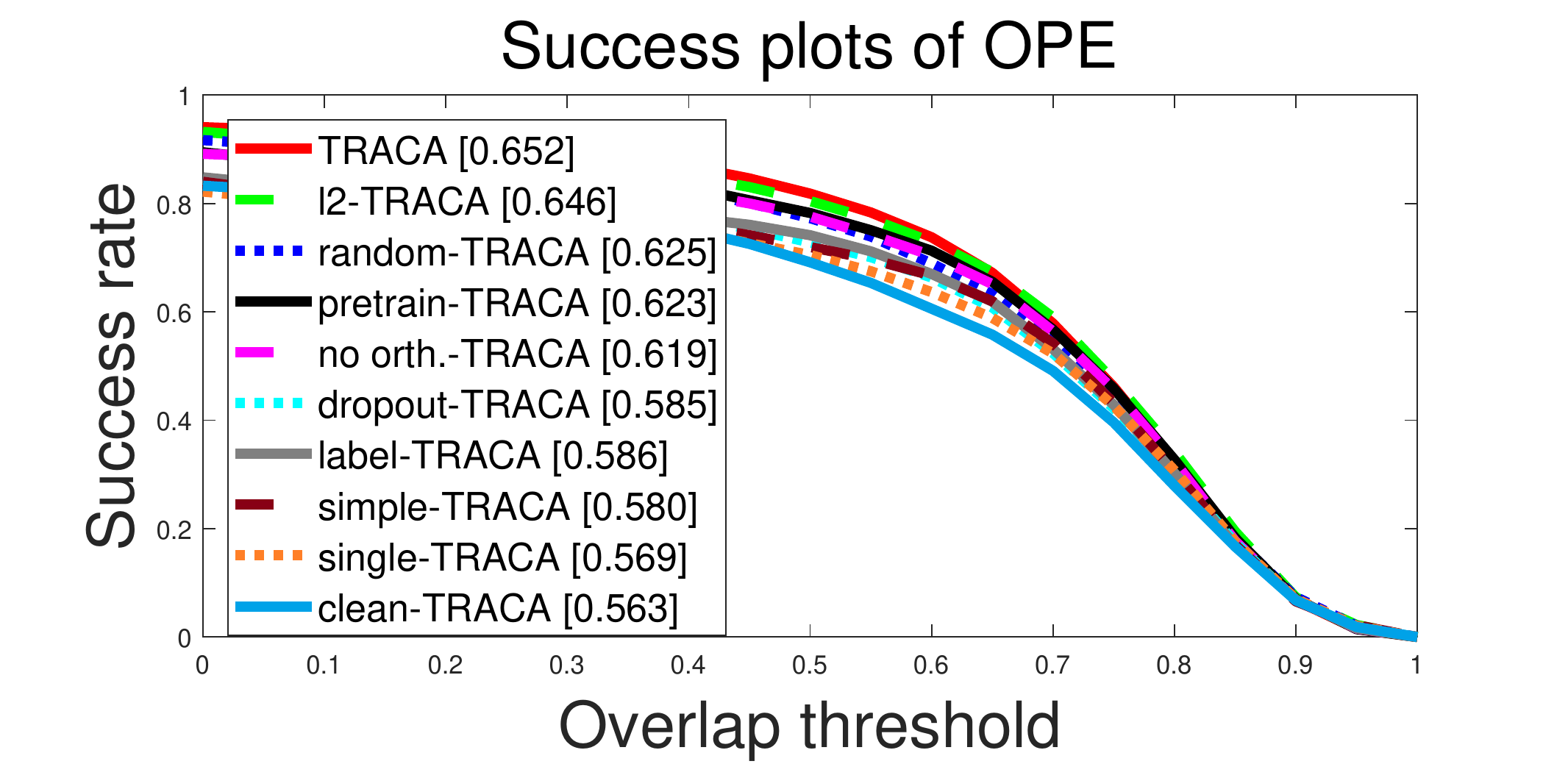}}
\hfill
\subfigure[Evaluation plots on CVPR2013]{\includegraphics[width=5.7cm]{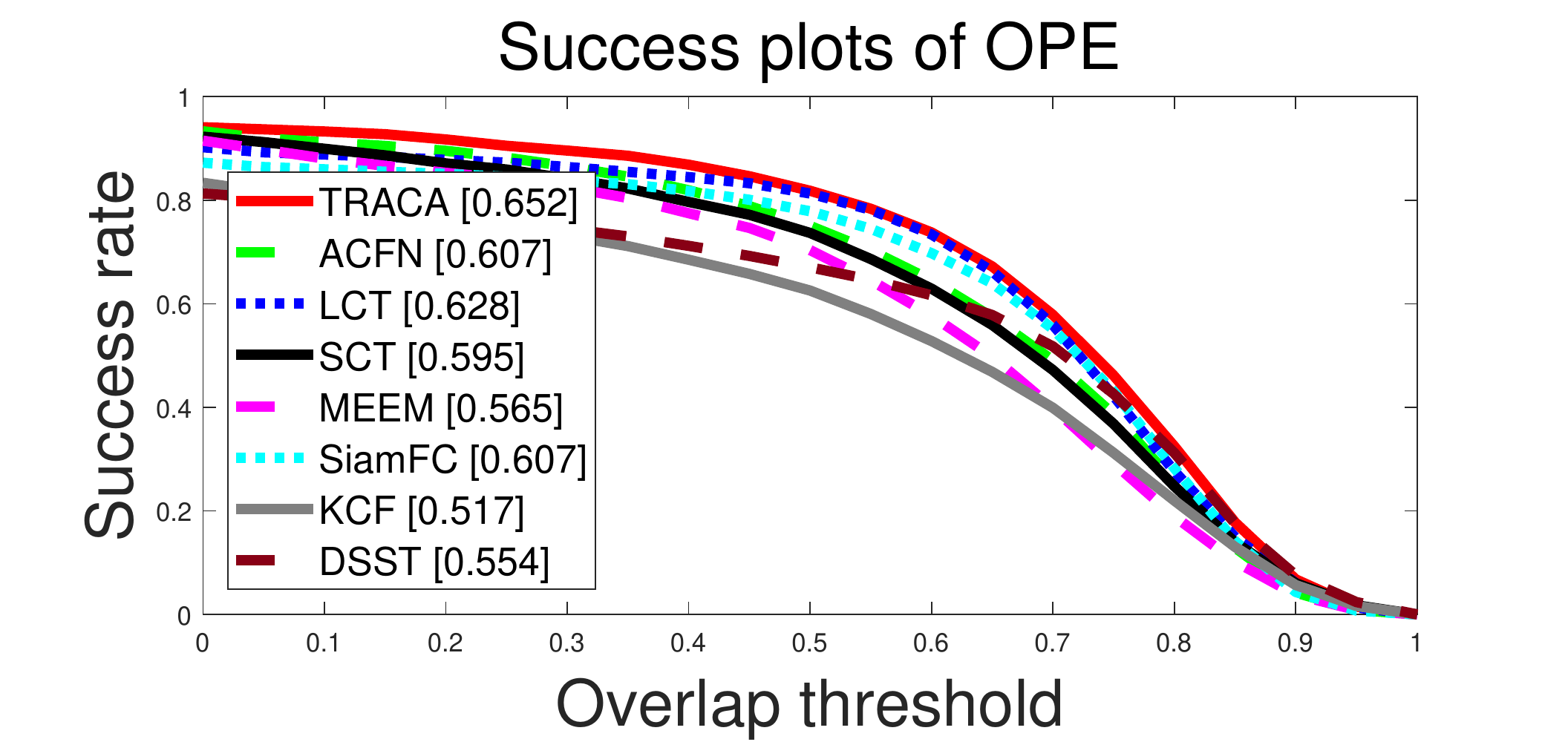}}
\hfill
\subfigure[Evaluation plots on TPAMI2015]{\includegraphics[width=5.7cm]{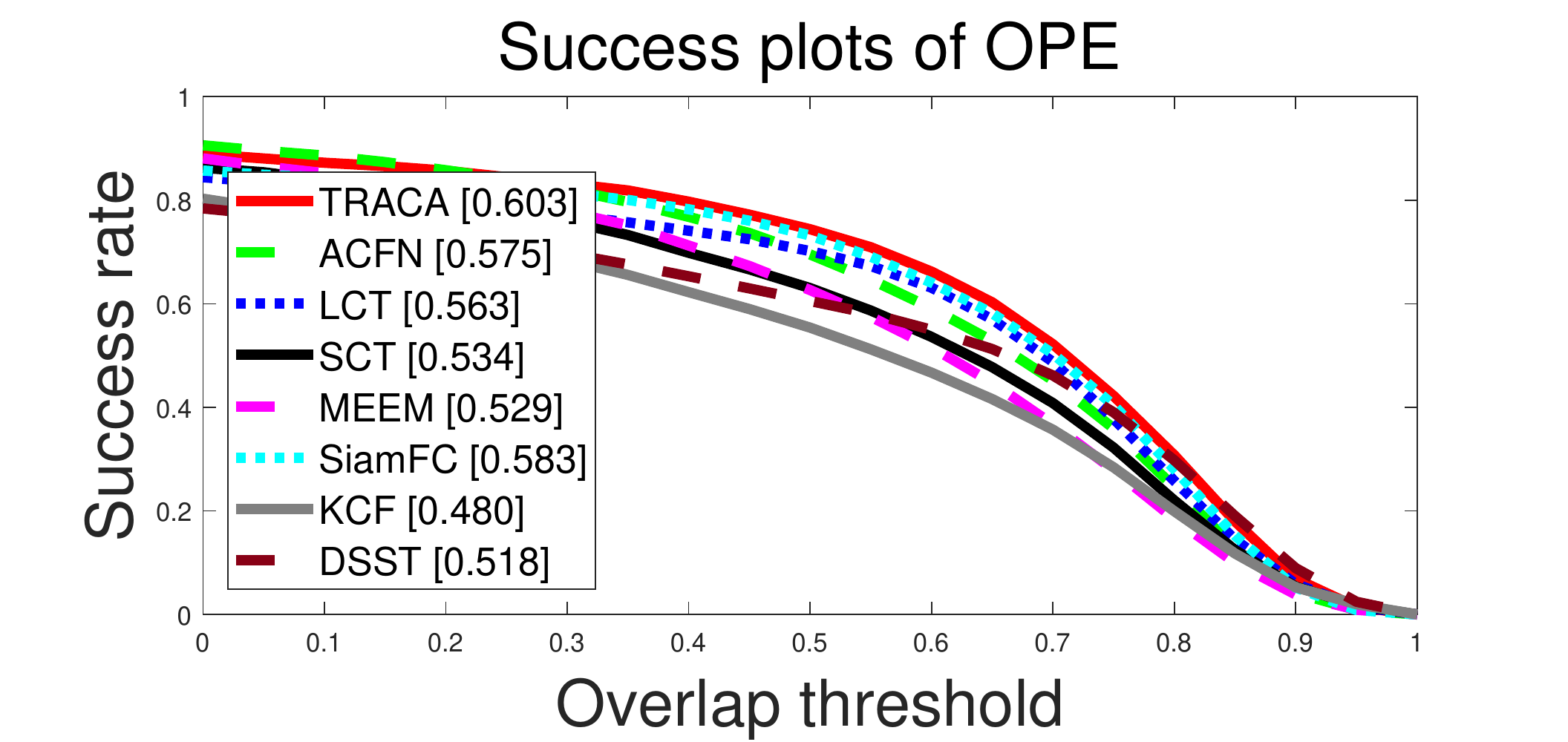}}
\caption{{\bf{Evaluation results.}} \ac{TRACA} showed the best performance within the self-comparison, and the state-of-the-art performance amongst real-time trackers in CVPR2013~\cite{ref:Benchmark} and TPAMI2015~\cite{ref:TPAMI2015Benchmark} datasets. The numbers within the legend are the average precisions when the centre error threshold equals 20 pixels (top row), or the area under the curve of the success plot (bottom row).}
    \label{fig:Evaluation}
    \vspace{-4mm}
\end{figure*}

\subsection{Dataset}
\label{subsec:dataset}
The classification image samples included in VOC2012~\cite{ref:voc2012} were used to pre-train the expert auto-encoders and the context-aware network.
To evaluate the proposed framework, we used the CVPR2013~\cite{ref:Benchmark} (51 targets, 50 videos) and TPAMI2015~\cite{ref:TPAMI2015Benchmark} (100 targets, 98 videos) datasets, which contain the ground truth of the target bounding box at every frame.
These datasets have been frequently used~\cite{ref:COT,ref:MDNet,ref:KCF,ref:MUSTer,ref:MEEM,ref:STRUCK,ref:ACFN} as they include a large variety of environments to evaluate the general tracking performance.

\subsection{Evaluation Measure}
As performance measure, we used the average precision curve of one-pass evaluation (OPE) as proposed in~\cite{ref:Benchmark}.
The average precision curve was estimated by averaging the precision curves of all sequences, which was obtained using two sources: location error threshold and overlap threshold.
As representative scores of trackers, the average precisions when the location error threshold equals 20 pixels and the area under the curve of the success plot were used.

\begin{table}
\centering 
\caption{Quantitative results on the CVPR2013 dataset~\cite{ref:Benchmark}}
\small{
\resizebox{0.9\linewidth}{!}{
\begin{tabular} {|c||c|>{\centering}m{1.4cm}|c|c|}
\hline
& Algorithm & Pre. score & Mean FPS & GPU\\
\hline
\multirow{10}{0.4cm}{\rotatebox[origin=c]{90}{Proposed}}
 & \textbf{TRACA} & \textbf{89.8\%} & 101.3 & Y\\
 \cline{2-5}
 & $l_2$-TRACA & 87.7\% & 101.2 & Y\\
 \cline{2-5}
 & random-TRACA & 84.4\% & 99.5 & Y\\
 \cline{2-5}
 & pretrain-TRACA & 83.2\% & 98.8 & Y\\
 \cline{2-5}
 & no orth.-TRACA & 82.2\% & 101.2 & Y\\
 \cline{2-5}
 & dropout-TRACA & 78.5\% & 97.5 & Y\\
 \cline{2-5}
 & label-TRACA & 76.6\% & 97.2 & Y\\
 \cline{2-5}
 & simple-TRACA & 76.4\% & 94.1 & Y\\
 \cline{2-5}
 & single-TRACA & 76.2\% & 100.9 & Y\\
 \cline{2-5}
 & clean-TRACA & 75.4\% & 92.9 & Y\\
\hline
\multirow{7}{0.4cm}{\rotatebox[origin=c]{90}{Real-time}}
& ACFN~\cite{ref:ACFN} & 86.0\% &  15.0 & Y\\
\cline{2-5}
& LCT~\cite{ref:LongCT} & 84.8\% &  21.6 & N\\
\cline{2-5}
& SCT~\cite{ref:SCT} & 84.5\% &  40.0 & N\\
\cline{2-5}
& MEEM~\cite{ref:MEEM} & 81.4\% & 19.5 & N\\
\cline{2-5}
& SiamFC~\cite{ref:SiamFC} & 80.9\% & 86.0 & Y\\
\cline{2-5}
& KCF~\cite{ref:KCF} & 74.2\% & \textbf{120.5} & N\\
\cline{2-5}
& DSST~\cite{ref:DSST} & 74.0\% & 25.4 & N\\
\hline
\hline
\multirow{6}{0.4cm}{\rotatebox[origin=c]{90}{Non Real-time}}
& ECO~\cite{ref:ECO} & {\bf{93.0\%}} & \textbf{8.0} & Y\\
\cline{2-5}
& ADNet~\cite{ref:yun} & 90.3\% & 2.9 & Y\\
\cline{2-5}
& C-COT~\cite{ref:COT} & 89.9\% & 0.5 & N\\
\cline{2-5}
& MUSTer~\cite{ref:MUSTer} & 86.5\% & 3.9 & N\\
\cline{2-5}
& FCNT~\cite{ref:FCNT} & 85.6\% & 3.0 & Y\\
\cline{2-5}
& D-SRDCF~\cite{ref:DeepSRDCF} & 84.9\% & 0.2 & N\\
\hline
\end{tabular}
}
\label{tabular:quantitative2}
}
\vspace{-3mm}
\end{table}

\subsection{Self-comparison}
To analyse the effectiveness of \ac{TRACA},
 we compare TRACA with nine variants:
no orth.-TRACA, pretrain-TRACA, clean-TRACA, dropout-TRACA, random-TRACA, $l_2$-TRACA, label-TRACA, simple-TRACA, and single-TRACA.
In no orth.-TRACA, the weight factor $\lambda_{\Theta}$ for the orthogonality loss term is set to zero.
Pretrain-TRACA skipped the initial adaptation step and directly utilised the pre-trained expert auto-encoder.
Clean-TRACA used the expert auto-encoders which were pre-trained without any extrinsic denoising process.
Dropout-TRACA adopted a dropout scheme instead of the proposed dimension corrupting process, while keeping the feature vector exchange process.
Random-TRACA randomly selected the suitable expert auto-encoder.
$l_2$-TRACA selected the best suitable expert auto-encoder according to the smallest $l_2$ generation error estimated from the initial target.
Label-TRACA utilised 20 class labels of the pre-training dataset (VOC2012~\cite{ref:voc2012}) as the contextual clusters.
The expert auto-encoders of simple-TRACA were trained and fine-tuned by minimising the Euclidean distance between their inputs and final outputs, \ie\ without using the multi-stage distance.
Single-TRACA utilised the compressed feature map of the base auto-encoder\footnote{For fair comparison, we train the base auto-encoder for 20 epochs in the case of single-TRACA.}.

The results of the comparison with these nine trackers are shown in Table~\ref{tabular:quantitative2} and Fig.~\ref{fig:Evaluation}~(a).
The results of random-TRACA and $l_2$-TRACA show decreased performance which reflects the importance of the context-aware network.
In the result of pretrain-TRACA, the performance was reduced by 6.6\% when the expert auto-encoder was not adapted initially.
The initial adaptation ignoring the orthogonality loss term (no orth.-TRACA) further decreased the performance by 1\% compared to pretrain-TRACA.
When the extrinsic denoising processes were not applied, the tracking performance reduced dramatically (14.3\%) according to the result of clean-TRACA.
Similarly, as shown in the result of dropout-TRACA, the proposed dimension corrupting process made the expert auto-encoders more robust than a dropout scheme (11.3\% performance reduction).
When the multi-stage distance was not used, the performance was reduced by 13.4\% as shown in the result of simple-TRACA.
Single-TRACA showed a dramatic reduction in the tracking performance (13.6\%), which demonstrates that the multiple-context scheme was effective to compress the raw deep convolutional feature for a specific target.
Finally, the tracking performance was reduced dramatically in label-TRACA (13.2\%), which shows that clustering in feature space is beneficial when training the expert auto-encoders.

\begin{figure}[t]
\centering
    \includegraphics[width=0.92\linewidth]{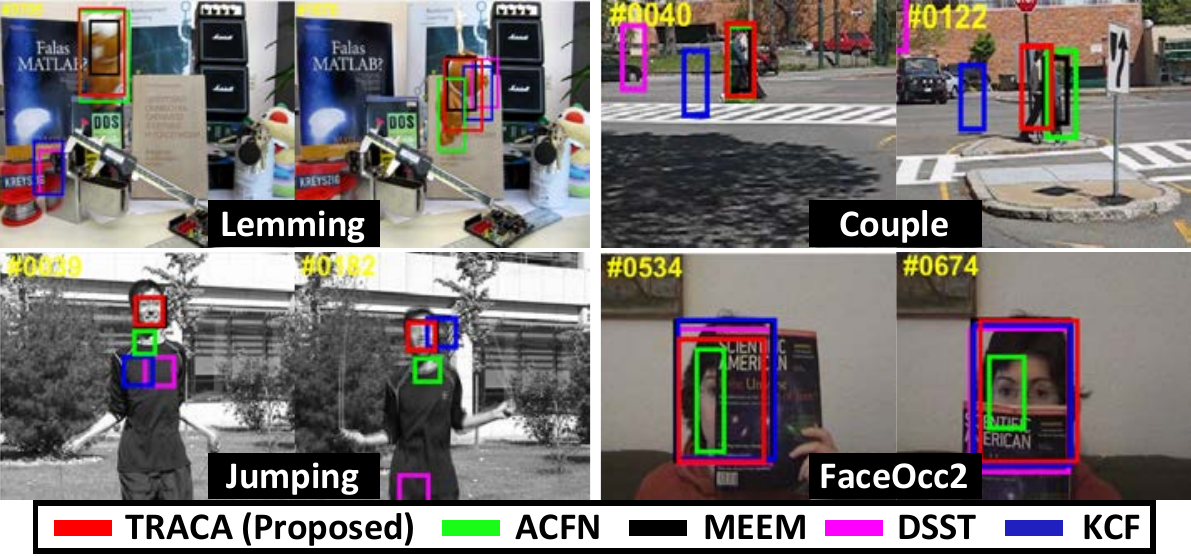}
    \caption{ {\bf{Qualitative results.}}  The used sequences are {{Lemming}}, {{Couple}}, {{Jumping}}, {{FaceOcc2}}, {{CarDark}}, and {{Soccer}} from the left-top.}
    \label{fig:Qualitative}    
    \vspace{-4mm}
\end{figure}

\subsection{Comparison with State-of-the-art Trackers}
The results of the state-of-the-art methods, including ECO~\cite{ref:ECO}, ADNet~\cite{ref:yun}, ACFN~\cite{ref:ACFN}, C-COT~\cite{ref:COT}, SiamFC~\cite{ref:SiamFC}, FCNT~\cite{ref:FCNT}, D-SRDCF~\cite{ref:DeepSRDCF}, SCT~\cite{ref:SCT}, LCT~\cite{ref:LongCT}, and DSST~\cite{ref:DSST} were obtained from the authors.
In addition, the results of MUSTer~\cite{ref:MUSTer}, MEEM~\cite{ref:MEEM}, and KCF~\cite{ref:KCF} were estimated using the authors' implementations\footnote{For fair comparison, the computational time was estimated by the same computer as \ac{TRACA} and included the image resizing time.}.

In Table~\ref{tabular:quantitative2}, the precision scores of the algorithms on the CVPR2013 dataset are presented along with the computational speed and whether they make use of a GPU.
%
\mbox{Fig.~\ref{fig:Evaluation}~(b-c)} compares the performances of the real-time trackers, where \ac{TRACA} demonstrates state-of-the-art performance in both the CVPR2013 and TPAMI2015 datasets while running at over 100 fps.
Some qualitative results are shown in Fig.~\ref{fig:Qualitative}.

\begin{figure}[t]
    \centering
\includegraphics[width=0.97\linewidth]{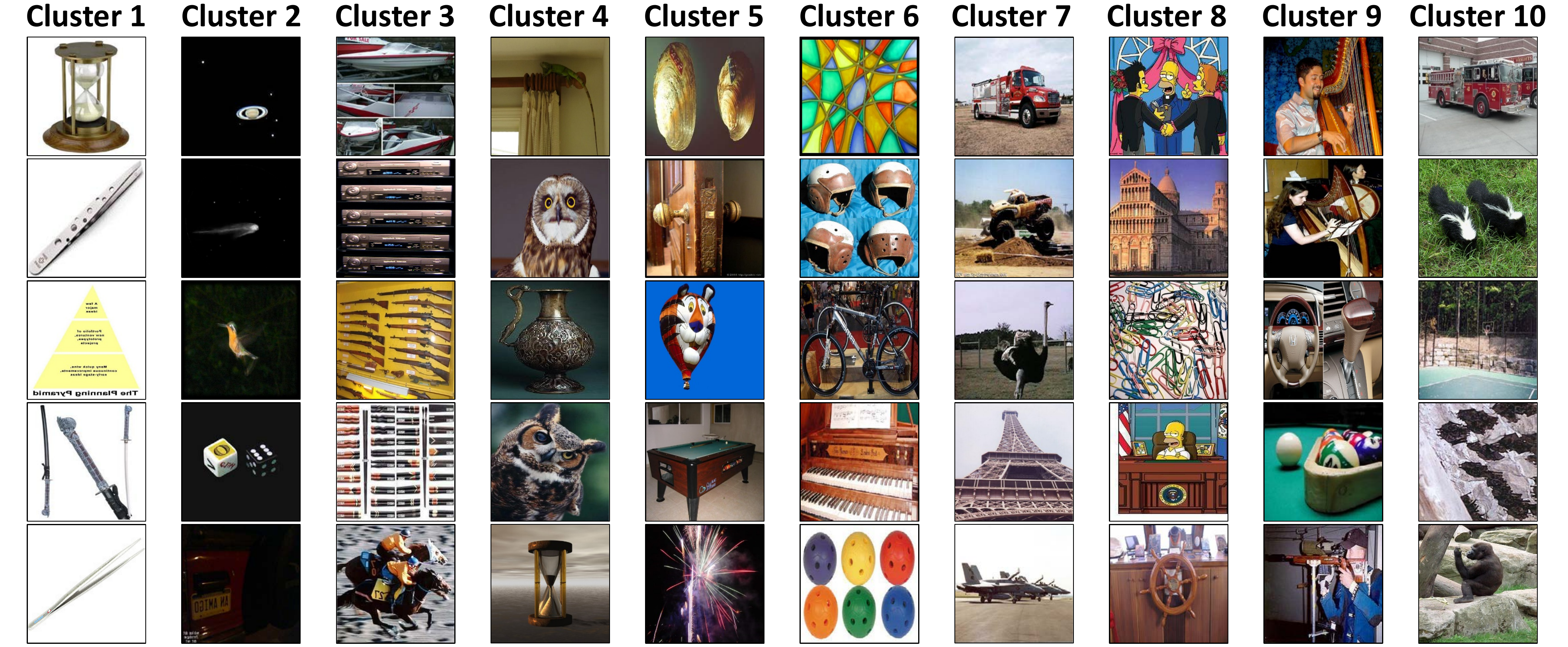}
	\caption{\textbf{Top-5 images for each contextual cluster.} Each column represents one context category and consists of the five samples within Caltech256~\cite{ref:caltech2} that have the highest scores of the context-aware network for this category. The results confirm that the contextual clusters represent the category of appearance patterns.}
	\label{fig:contextClutering}
    \vspace{-5mm}
\end{figure}

\subsection{Further Analysis}
The context in the proposed framework refers to a coarse category of the compressed feature maps encoding the target object appearance.
To illustrate the context in practice, we extracted the five samples with the highest scores within the context-aware network for each contextual category using Caltech256~\cite{ref:caltech2}.
As shown in Fig.~\ref{fig:contextClutering}, the contextual clusters categorise the samples according to appearance patterns.

In Appendix B, we evaluate the impact of the chosen target layer of VGG-Net and the number of contextual clusters on the proposed framework.
In Appendix C, we analyse the correlation matrix among various computer vision datasets, which was obtained by estimating the correlation among the histograms of the results from the context-aware network.

\vspace{-2mm}
\section{Conclusion}
\vspace{-3mm}
In this paper, a new visual tracking framework based on context-aware deep feature compression using multiple auto-encoders was proposed. 
Our main contribution is to introduce a context-aware scheme which includes expert auto-encoders specialising in one context, and a context-aware network which is able to select the best expert auto-encoder for a specific tracking target.
This leads to a significant speed-up of the correlation filter based tracker utilising deep convolutional features. Our experiments lead to the compelling finding that our framework achieves a high-speed tracking ability of over 100 fps while our framework maintains a competitive performance compared to the state-of-the-art trackers. We expect that embedding our context-aware deep feature compression scheme will be integrated with other trackers utilising raw deep features, which will increase their robustness and computational efficiency.
In addition, the scheme can be utilised as a way to avoid the overfitting problem in other computer vision tasks where only few training samples are available, such as in image $k$-shot learning and image domain adaptation.
As a future work, we will jointly train the expert auto-encoders and the context-aware network to potentially further increase the performance due to the correlation between the contextual clustering and the feature compression.

\vspace{1mm}
\footnotesize
\noindent\textbf{Acknowledgements:} 
This work was supported by  ICT R\&D program MSIP/IITP [2017-0-00306, Outdoor Surveillance Robots], Next-Generation ICD Program through NRF funded by Ministry of S\&ICT [2017M3C4A7077582], 
and BK21$^+$.
\normalsize

{\small
\bibliographystyle{ieee}
\bibliography{refs}

\begin{thebibliography}{10}\itemsep=-1pt

\bibitem{ref:SiamFC}
L.~Bertinetto, J.~Valmadre, J.~F. Henriques, A.~Vedaldi, and P.~H. Torr.
\newblock Fully-convolutional siamese networks for object tracking.
\newblock In {\em ECCV workshop}, 2016.

\bibitem{ref:MOSSE}
D.~S. Bolme, J.~R. Beveridge, B.~A. Draper, and Y.~M. Lui.
\newblock Visual object tracking using adaptive correlation filters.
\newblock In {\em CVPR}, 2010.

\bibitem{ref:vggm}
K.~Chatfield, K.~Simonyan, A.~Vedaldi, and A.~Zisserman.
\newblock Return of the devil in the details: Delving deep into convolutional
  nets.
\newblock In {\em BMVC}, 2014.

\bibitem{ref:SCT}
J.~Choi, H.~J. Chang, J.~Jeong, Y.~Demiris, and J.~Y. Choi.
\newblock Visual tracking using attention-modulated disintegration and
  integration.
\newblock In {\em CVPR}, 2016.

\bibitem{ref:ACFN}
J.~Choi, H.~J. Chang, S.~Yun, T.~Fischer, Y.~Demiris, and J.~Y. Choi.
\newblock Attentional correlation filter network for adaptive visual tracking.
\newblock In {\em CVPR}, 2017.

\bibitem{ref:ECO}
M.~Danelljan, G.~Bhat, F.~S. Khan, and M.~Felsberg.
\newblock {ECO:} efficient convolution operators for tracking.
\newblock In {\em CVPR}, 2017.

\bibitem{ref:DeepSRDCF}
M.~Danelljan, G.~Hager, F.~S. Khan, and M.~Felsberg.
\newblock Convolutional features for correlation filter based visual tracking.
\newblock In {\em ICCV workshops}, 2016.

\bibitem{ref:DSST}
M.~Danelljan, G.~Hager, F.~S. Khan, and M.~Felsberg.
\newblock Discriminative scale space tracking.
\newblock {\em IEEE Trans. on PAMI}, 39(8):1561--1575, 2016.

\bibitem{ref:SRDCF}
M.~Danelljan, G.~Häger, F.~Khan, and M.~Felsberg.
\newblock Learning spatially regularized correlation filters for visual
  tracking.
\newblock In {\em ICCV}, 2015.

\bibitem{ref:COT}
M.~Danelljan, A.~Robinson, F.~S. Khan, and M.~Felsberg.
\newblock Beyond correlation filters: Learning continuous convolution operators
  for visual tracking.
\newblock In {\em ECCV}, 2016.

\bibitem{ref:ae3}
B.~Du, W.~Xiong, J.~Wu, L.~Zhang, L.~Zhang, and D.~Tao.
\newblock Stacked convolutional denoising auto-encoders for feature
  representation.
\newblock {\em IEEE Trans. on Cybernetics}, 47(4):1017--1027, 2017.

\bibitem{ref:voc2012}
M.~Everingham, L.~Van~Gool, C.~K.~I. Williams, J.~Winn, and A.~Zisserman.
\newblock The {PASCAL} {V}isual {O}bject {C}lasses {C}hallenge 2012 {(VOC2012)}
  {R}esults.
\newblock
  http://www.pascal-network.org/challenges/VOC/voc2012/workshop/index.html.

\bibitem{ref:caltech2}
G.~Griffin, A.~Holub, and P.~Perona.
\newblock Caltech-256 object category dataset.
\newblock In {\em Caltech Technical Report}. California Institute of
  Technology, 2007.

\bibitem{ref:STRUCK}
S.~Hare, S.~Golodetz, A.~Saffari, V.~Vineet, M.~M. Cheng, S.~L. Hicks, and
  P.~H.~S. Torr.
\newblock Struck: Structured output tracking with kernels.
\newblock {\em IEEE Trans. on PAMI}, 38(10):2096--2109, 2016.

\bibitem{ref:goturn}
D.~Held, S.~Thrun, and S.~Savarese.
\newblock Learning to track at 100 fps with deep regression networks.
\newblock In {\em ECCV}, 2016.

\bibitem{ref:KCF}
J.~F. Henriques, R.~Caseiro, P.~Martins, and J.~Batista.
\newblock High-speed tracking with kernelized correlation filters.
\newblock {\em IEEE Trans. on PAMI}, 37(3):583--596, 2015.

\bibitem{ref:ae_hinton2006_2}
G.~E. Hinton, S.~Osindero, and Y.-W. Teh.
\newblock A fast learning algorithm for deep belief nets.
\newblock {\em Neural Computation}, 18(7):1527--1554, 2006.

\bibitem{ref:ae_hinton2006}
G.~E. Hinton and R.~R. Salakhutdinov.
\newblock Reducing the dimensionality of data with neural networks.
\newblock {\em Science}, 313(5786):504--507, 2006.

\bibitem{ref:MUSTer}
Z.~Hong, Z.~Chen, C.~Wang, X.~Mei, D.~Prokhorov, and D.~Tao.
\newblock Multi-store tracker ({MUSTer}): a cognitive psychology inspired
  approach to object tracking.
\newblock In {\em CVPR}, 2015.

\bibitem{ref:context_cascadeface}
H.~Li, Z.~Lin, X.~Shen, J.~Brandt, and G.~Hua.
\newblock A convolutional neural network cascade for face detection.
\newblock In {\em CVPR}, 2015.

\bibitem{ref:CF2}
C.~Ma, J.-B. Huang, X.~Yang, and M.-H. Yang.
\newblock Hierarchical convolutional features for visual tracking.
\newblock In {\em ICCV}, 2015.

\bibitem{ref:LongCT}
C.~Ma, X.~Yang, C.~Zhang, and M.-H. Yang.
\newblock Long-term correlation tracking.
\newblock In {\em CVPR}, 2015.

\bibitem{ref:ae2}
R.~Memisevic.
\newblock Gradient-based learning of higher-order image features.
\newblock In {\em ICCV}, 2011.

\bibitem{ref:MDNet}
H.~Nam and B.~Han.
\newblock Learning multi-domain convolutional neural networks for visual
  tracking.
\newblock In {\em CVPR}, 2016.

\bibitem{ref:crowd2}
D.~O{\~{n}}oro-Rubio and R.~J. L{\'o}pez-Sastre.
\newblock Towards perspective-free object counting with deep learning.
\newblock In {\em ECCV}, 2016.

\bibitem{ref:HDT}
Y.~Qi, S.~Zhang, L.~Qin, H.~Yao, Q.~Huang, J.~Lim, and M.-H. Yang.
\newblock Hedged deep tracking.
\newblock In {\em CVPR}, 2016.

\bibitem{ref:ImageNet}
O.~Russakovsky, J.~Deng, H.~Su, J.~Krause, S.~Satheesh, S.~Ma, Z.~Huang,
  A.~Karpathy, A.~Khosla, M.~Bernstein, A.~C. Berg, and L.~Fei-Fei.
\newblock Imagenet large scale visual recognition challenge.
\newblock {\em IJCV}, 115(3):211--252, 2015.

\bibitem{ref:crowd3}
D.~B. Sam, S.~Surya, and R.~V. Babu.
\newblock Switching convolutional neural network for crowd counting.
\newblock In {\em CVPR}, 2017.

\bibitem{ref:SINT}
R.~Tao, E.~Gavves, and A.~W. Smeulders.
\newblock Siamese instance search for tracking.
\newblock In {\em CVPR}, 2016.

\bibitem{ref:matconvnet}
A.~Vedaldi and K.~Lenc.
\newblock Matconvnet -- convolutional neural networks for matlab.
\newblock In {\em ACM MM}, 2015.

\bibitem{ref:dAE}
P.~Vincent, H.~Larochelle, Y.~Bengio, and P.-A. Manzagol.
\newblock Extracting and composing robust features with denoising autoencoders.
\newblock In {\em ICML}, 2008.

\bibitem{ref:context_head}
T.-H. Vu, A.~Osokin, and I.~Laptev.
\newblock Context-aware {CNNs} for person head detection.
\newblock In {\em ICCV}, 2015.

\bibitem{ref:context_saliency2}
L.~Wang, H.~Lu, X.~Ruan, and M.-H. Yang.
\newblock Deep networks for saliency detection via local estimation and global
  search.
\newblock In {\em CVPR}, 2015.

\bibitem{ref:FCNT}
L.~Wang, W.~Ouyang, X.~Wang, and H.~Lu.
\newblock Visual tracking with fully convolutional networks.
\newblock In {\em ICCV}, 2015.

\bibitem{ref:STCT}
L.~Wang, W.~Ouyang, X.~Wang, and H.~Lu.
\newblock {STCT}: Sequentially training convolutional networks for visual
  tracking.
\newblock In {\em CVPR}, 2016.

\bibitem{ref:Benchmark}
Y.~Wu, J.~Lim, and M.-H. Yang.
\newblock Online object tracking: A benchmark.
\newblock In {\em CVPR}, 2013.

\bibitem{ref:TPAMI2015Benchmark}
Y.~Wu, J.~Lim, and M.-H. Yang.
\newblock Object tracking benchmark.
\newblock {\em IEEE Trans. on PAMI}, 37(9):1834--1848, 2015.

\bibitem{ref:ae1}
X.~Yan, J.~Yang, K.~Sohn, and H.~Lee.
\newblock Attribute2image: Conditional image generation from visual attributes.
\newblock In {\em ECCV}, 2016.

\bibitem{ref:yun}
S.~Yun, J.~Choi, Y.~Yoo, K.~Yun, and J.~Y. Choi.
\newblock Action-decision networks for visual tracking with deep reinforcement
  learning.
\newblock In {\em CVPR}, 2017.

\bibitem{ref:MEEM}
J.~Zhang, S.~Ma, and S.~Sclaroff.
\newblock Meem: Robust tracking via multiple experts using entropy
  minimization.
\newblock In {\em ECCV}, 2014.

\bibitem{ref:crowd1}
Y.~Zhang, D.~Zhou, S.~Chen, S.~Gao, and Y.~Ma.
\newblock Single-image crowd counting via multi-column convolutional neural
  network.
\newblock In {\em CVPR}, 2016.

\bibitem{ref:context_saliency}
R.~Zhao, W.~Ouyang, H.~Li, and X.~Wang.
\newblock Saliency detection by multi-context deep learning.
\newblock In {\em CVPR}, 2015.

\end{thebibliography}


@inproceedings{ref:goturn,
  title={Learning to Track at 100 FPS with Deep Regression Networks},
  author={Held, David and Thrun, Sebastian and Savarese, Silvio},
  booktitle = {ECCV},
  year      = {2016}
}

@inproceedings{ref:ae1,
author="Yan, Xinchen and Yang, Jimei and Sohn, Kihyuk and Lee, Honglak",
title="Attribute2Image: Conditional Image Generation from Visual Attributes",
bookTitle="ECCV",
year="2016",
}

@INPROCEEDINGS{ref:ae2, 
author={R. Memisevic}, 
booktitle={ICCV}, 
title={Gradient-based learning of higher-order image features}, 
year={2011}, 
}

@ARTICLE{ref:ae3, 
author={B. Du and W. Xiong and J. Wu and L. Zhang and L. Zhang and D. Tao}, 
journal={IEEE Trans. on Cybernetics}, 
title={Stacked Convolutional Denoising Auto-Encoders for Feature Representation}, 
year={2017}, 
volume={47}, 
number={4}, 
pages={1017-1027}, 
}



@inproceedings{ref:BACF,
author = {Kiani Galoogahi, Hamed and Fagg, Ashton and Lucey, Simon},
title = {Learning Background-Aware Correlation Filters for Visual Tracking},
  booktitle = {ICCV},
	year = {2017},
}

@inproceedings{ref:uav,
  title={A Benchmark and Simulator for UAV Tracking},
 author={Matthias Mueller and Neil Smith and Bernard Ghanem},
  booktitle = {ECCV},
	year = {2016},
}

@article{ref:caltech1,
	author="Fei-Fei, Li and Fergus, Rob and Perona, Pietro",
	title="Learning generative visual models from few training examples: An incremental bayesian approach tested on 101 object categories",
	journal="CVIU",
	year="2007",
	volume="106",
	number="1",
	pages="59--70"
}

@inproceedings{ref:caltech2,
  title={Caltech-256 object category dataset},
  author={Griffin, Gregory and Holub, Alex and Perona, Pietro},
  year={2007},
  booktitle={Caltech Technical Report},
  publisher={California Institute of Technology}
}

@inproceedings{ref:glorot2011,
  title={Domain adaptation for large-scale sentiment classification: A deep learning approach},
  author={Glorot, Xavier and Bordes, Antoine and Bengio, Yoshua},
  booktitle={ICML},
  year={2011}
}

@inproceedings{ref:masci2011,
author="Masci, Jonathan and Meier, Ueli and Cire{\c{s}}an, Dan and Schmidhuber, J{\"u}rgen",
title="Stacked Convolutional Auto-Encoders for Hierarchical Feature Extraction",
bookTitle="ICANN",
year="2011",
}



@inproceedings{ref:chen,
 author = {Chen, Minmin and Xu, Zhixiang and Weinberger, Kilian Q. and Sha, Fei},
 title = {Marginalized Denoising Autoencoders for Domain Adaptation},
 booktitle = {ICML},
 year = {2012},
} 

@article{ref:ae_hinton2006_2,
  title={A fast learning algorithm for deep belief nets},
  author={Hinton, Geoffrey E and Osindero, Simon and Teh, Yee-Whye},
  journal={Neural Computation},
  volume={18},
  number={7},
  pages={1527--1554},
  year={2006},
}

@article{ref:ae_hinton2006,
  title={Reducing the dimensionality of data with neural networks},
  author={Hinton, Geoffrey E and Salakhutdinov, Ruslan R},
  journal={Science},
  volume={313},
  number={5786},
  pages={504--507},
  year={2006},
}


@inproceedings{ref:arxiv_comp2,
  author    = {Yong{-}Deok Kim and
               Eunhyeok Park and
               Sungjoo Yoo and
               Taelim Choi and
               Lu Yang and
               Dongjun Shin},
  title     = {Compression of Deep Convolutional Neural Networks for Fast and Low
               Power Mobile Applications},
    booktitle = {arXiv},
	year = {2015},
    pages = {}
}

@inproceedings{ref:arxiv_comp1,
  author    = {Yunchao Gong and
               Liu Liu and
               Ming Yang and
               Lubomir D. Bourdev},
  title     = {Compressing Deep Convolutional Networks using Vector Quantization},
    booktitle = {arXiv},
	year = {2014},
    pages = {}
}


@article{ref:prior2,
  title={Modeling the effects of prior knowledge on learning incongruent features of category members.},
  author={Heit, Evan and Briggs, Janet and Bott, Lewis},
  journal={Journal of Experimental Psychology: Learning, Memory, and Cognition},
  volume={30},
  number={5},
  pages={1065},
  year={2004},
  publisher={American Psychological Association}
}


@incollection{ref:cate1,
title = "Knowledge selection in category learning",
series = "Psychology of Learning and Motivation",
publisher = "Academic Press",
volume = "39",
number = "Supplement C",
pages = "163 - 199",
year = "2000",
author = "Evan Heit and Lewis Bott"
}


@article{ref:prior1,
title = "Explanation and prior knowledge interact to guide learning",
journal = "Cognitive Psychology",
volume = "66",
number = "1",
pages = "55 - 84",
year = "2013",
author = "Joseph J. Williams and Tania Lombrozo",
}


 @inproceedings{ref:matconvnet,
      author    = {A. Vedaldi and K. Lenc},
      title     = {MatConvNet -- Convolutional Neural Networks for MATLAB},
      booktitle = {ACM MM},
      year      = {2015},
  }
  

@inproceedings{ref:vggm,
  author    = {Ken Chatfield and
               Karen Simonyan and
               Andrea Vedaldi and
               Andrew Zisserman},
  title     = {Return of the Devil in the Details: Delving Deep into Convolutional
               Nets},
  booktitle = {BMVC}, 
  year      = {2014},
}

@misc{ref:voc2012,
	author = "Everingham, M. and Van~Gool, L. and Williams, C. K. I. and Winn, J. and Zisserman, A.",
	title = "The {PASCAL} {V}isual {O}bject {C}lasses {C}hallenge 2012 {(VOC2012)} {R}esults",
	howpublished = "http://www.pascal-network.org/challenges/VOC/voc2012/workshop/index.html"}
    
@inproceedings{ref:dAE,
 author = {Vincent, Pascal and Larochelle, Hugo and Bengio, Yoshua and Manzagol, Pierre-Antoine},
 title = {Extracting and Composing Robust Features with Denoising Autoencoders},
 booktitle = {ICML},
 year = {2008},
} 

@INPROCEEDINGS{ref:comp3,
  author    = {Tien{-}Ju Yang and
               Yu{-}Hsin Chen and
               Vivienne Sze},
  title     = {Designing Energy-Efficient Convolutional Neural Networks using Energy-Aware Pruning},
  booktitle = {CVPR},
  year = {2017},
  }
  
@INPROCEEDINGS{ref:comp2,
  author    = {Hao Li and
               Asim Kadav and
               Igor Durdanovic and
               Hanan Samet and
               Hans Peter Graf},
  title     = {Pruning Filters for Efficient ConvNets},
  booktitle={ICLR}, 
  year      = {2017},
}

@INPROCEEDINGS{ref:comp1,
  author    = {Song Han and
               Huizi Mao and
               William J. Dally},
  title     = {Deep Compression: Compressing Deep Neural Network with Pruning, Trained
               Quantization and Huffman Coding},
  booktitle={ICLR}, 
  year      = {2016},
}

@inproceedings{ref:crowd1,
  title={Single-image crowd counting via multi-column convolutional neural network},
  author={Zhang, Yingying and Zhou, Desen and Chen, Siqin and Gao, Shenghua and Ma, Yi},
  booktitle={CVPR},
  year={2016}
}

@inproceedings{ref:crowd2,
author="O{\~{n}}oro-Rubio, Daniel
and L{\'o}pez-Sastre, Roberto J.",
title="Towards Perspective-Free Object Counting with Deep Learning",
bookTitle="ECCV",
year="2016",
}

@inproceedings{ref:crowd3,
  title={Switching Convolutional Neural Network for Crowd Counting},
  author={Deepak Babu Sam and Shiv Surya and R. Venkatesh Babu},
  booktitle={CVPR},  
  year={2017}
}



@inproceedings{ref:context_cascadeface,
  title={A convolutional neural network cascade for face detection},
  author={Li, Haoxiang and Lin, Zhe and Shen, Xiaohui and Brandt, Jonathan and Hua, Gang},
  booktitle={CVPR},
  year={2015}
}


@InProceedings{ref:context_head,
author = {Vu, Tuan-Hung and Osokin, Anton and Laptev, Ivan},
title = {Context-Aware {CNNs} for Person Head Detection},
booktitle = {ICCV},
year = {2015}
}

@InProceedings{ref:context_detection,
author="Kantorov, Vadim
and Oquab, Maxime
and Cho, Minsu
and Laptev, Ivan",
title="ContextLocNet: Context-Aware Deep Network Models for Weakly Supervised Localization",
booktitle="ECCV",
year = "2016",
}

@InProceedings{ref:context_saliency,
author = {Zhao, Rui and Ouyang, Wanli and Li, Hongsheng and Wang, Xiaogang},
title = {Saliency Detection by Multi-Context Deep Learning},
booktitle = {CVPR},
year = {2015}
}

@INPROCEEDINGS{ref:context_speech4, 
author={L. Deng and J. Li and J. T. Huang and K. Yao and D. Yu and F. Seide and M. Seltzer and G. Zweig and X. He and J. Williams and Y. Gong and A. Acero}, 
booktitle={ICASSP}, 
title={Recent advances in deep learning for speech research at Microsoft}, 
year={2013}, 
}

@INPROCEEDINGS{ref:context_speech3, 
author={H. Liao}, 
booktitle={ICASSP}, 
title={Speaker adaptation of context dependent deep neural networks}, 
year={2013}, 
}

@ARTICLE{ref:context_speech1, 
author={G. E. Dahl and D. Yu and L. Deng and A. Acero}, 
journal={IEEE Transactions on Audio, Speech, and Language Processing}, 
title={Context-Dependent Pre-Trained Deep Neural Networks for Large-Vocabulary Speech Recognition}, 
year={2012}, 
volume={20}, 
number={1}, 
pages={30-42}, }

@INPROCEEDINGS{ref:context_speech2, 
author={F. Seide and G. Li and X. Chen and D. Yu}, 
booktitle={IEEE Workshop on Automatic Speech Recognition Understanding}, 
title={Feature engineering in Context-Dependent Deep Neural Networks for conversational speech transcription}, 
year={2011}, 
}


@inproceedings{ref:DTTD,
  title={Detect to Track and Track to Detect },
  author={C. Feichtenhofer and A. Pinz and A. Zisserman},
  booktitle={ICCV},  
  year={2017}
}


@inproceedings{ref:multi_ales,
  author    = {Jingjing Xiao and
               Qiang Lan and
               Linbo Qiao and
               Ales Leonardis},
  title     = {Semantic tracking: Single-target tracking with inter-supervised convolutional
               networks},
    booktitle = {arXiv},
	year = {2016},
    pages = {}
}


@INPROCEEDINGS{ref:multi_nl2, 
author={L. Deng and J. Li and J. T. Huang and K. Yao and D. Yu and F. Seide and M. Seltzer and G. Zweig and X. He and J. Williams and Y. Gong and A. Acero}, 
booktitle={ICASSP}, 
title={Recent advances in deep learning for speech research at Microsoft}, 
year={2013}, 
}

@inproceedings{ref:multi_nl1,
 author = {Collobert, Ronan and Weston, Jason},
 title = {A Unified Architecture for Natural Language Processing: Deep Neural Networks with Multitask Learning},
 booktitle = {ICML},
 year = {2008},
} 

@INPROCEEDINGS{ref:multi_face2, 
author={C. Zhang and Z. Zhang}, 
booktitle={WACV}, 
title={Improving multiview face detection with multi-task deep convolutional neural networks}, 
year={2014}, 
}

@Inbook{ref:multi_face1,
author="Zhang, Zhanpeng
and Luo, Ping
and Loy, Chen Change
and Tang, Xiaoou",
title="Facial Landmark Detection by Deep Multi-task Learning",
bookTitle="ECCV",
year="2014",
}



@article{ref:multitask_yoshua,
url = {http://dx.doi.org/10.1561/2200000006},
year = {2009},
volume = {2},
journal = {Foundations and Trends® in Machine Learning},
title = {Learning Deep Architectures for AI},
doi = {10.1561/2200000006},
issn = {1935-8237},
number = {1},
pages = {1-127},
author = {Yoshua Bengio}
}


@inproceedings{ref:ACFN,
  title={Attentional Correlation Filter Network for Adaptive Visual Tracking},
  author={Choi, Jongwon and Chang, Hyung Jin and Yun, Sangdoo and Fischer, Tobias and Demiris, Yiannis and Choi, Jin Young},
  booktitle={CVPR},  
  year={2017}
}

@inproceedings{ref:ECO,
  title={{ECO:} Efficient Convolution Operators for Tracking},
  author={Martin Danelljan and
               Goutam Bhat and
               Fahad Shahbaz Khan and
               Michael Felsberg},
  booktitle={CVPR},  
  year={2017},
}

@article{ref:DSST, 
author={M. Danelljan and G. Hager and F. S. Khan and M. Felsberg}, 
journal={IEEE Trans. on PAMI}, 
title={Discriminative Scale Space Tracking}, 
volume={39},
number={8},
pages={1561-1575},
year=2016
}

@inproceedings{ref:bibi,
	author = {Bibi, Adel and Mueller, Matthias and Ghanem, Bernard},
    title = {Target Response Adaptation for Correlation Filter Tracking},
    booktitle = {ECCV},
    year = {2016}
}

@inproceedings{ref:yun,
  title={Action-Decision Networks for Visual Tracking with Deep Reinforcement Learning},
  author={Yun, Sangdoo and Choi, Jongwon and Yoo, Youngjoon and Yun, Kimin and Choi, Jin Young},
  booktitle={CVPR},  
  year={2017},
}

@article{ref:KCF,
	author = {João F Henriques and Rui Caseiro and Pedro Martins and Jorge Batista},
    title = {High-speed tracking with kernelized correlation filters},
    journal = {IEEE Trans. on PAMI},
	volume = {37},
	number = {3},
	year = {2015},
	pages = {583-596}    
}

@inproceedings{ref:MOSSE,
	author = {David S Bolme and J Ross Beveridge and Bruce A Draper and Yui Man Lui},
    title = {Visual object tracking using adaptive correlation filters},
    booktitle = {CVPR},
	year = {2010}
}

@article{ref:TLD,
	author = {Zdenek Kalal and Krystian Mikolajczyk and Jiri Matas},
    title = {Tracking-Learning-Detection},
    journal = {IEEE Trans. on PAMI},
	volume = {34},
	number = {7},
	year = {2012},
	pages = {1409-1422}    
}

@article{ref:STRUCK,
	author = {S. Hare and S. Golodetz and A. Saffari and V. Vineet and M. M. Cheng and S. L. Hicks and P. H. S. Torr},
    title={Struck: Structured Output Tracking with Kernels},
    journal = {IEEE Trans. on PAMI},
	volume = {38},
	number = {10},
	year = {2016},
	pages = {2096-2109}        
}

@inproceedings{ref:MUSTer,
	author = {Z. Hong and Zhe Chen and C. Wang and X. Mei and D. Prokhorov and D. Tao},
    title = {MUlti-Store Tracker ({MUSTer}): a Cognitive Psychology Inspired Approach to Object Tracking},
    booktitle = {CVPR},
	year = {2015}    
}

@InProceedings{ref:context_saliency2,
author = {Wang, Lijun and Lu, Huchuan and Ruan, Xiang and Yang, Ming-Hsuan},
title = {Deep Networks for Saliency Detection via Local Estimation and Global Search},
booktitle = {CVPR},
year = {2015}
}



@inproceedings{ref:MEEM,
	author = {J. Zhang and S. Ma and S. Sclaroff},
    title = {MEEM: Robust Tracking via Multiple Experts using Entropy Minimization},
    booktitle = {ECCV},
    year = {2014}
}

@inproceedings{ref:VTD,
	author = {J. Kwon and K. M. Lee},
	title = {Visual Tracking Decomposition},
	booktitle = {CVPR},
	year = {2010},
}

@inproceedings{ref:VTS,
	author = {J. Kwon and K. M. Lee},
	title = {Tracking by Sampling Trackers},
	booktitle = {ICCV},
	year = {2011},
}

@inproceedings{ref:SCM,
	author = {W. Zhong and H. Lu and M.-H. Yang},
	title = {Robust Object Tracking via Sparsity-based Collaborative Model},
	booktitle = {CVPR},
	year = {2012},
}

@inproceedings{ref:ASLA,
	author = {X. Jia and H. Lu and M.-H. Yang},
	title = {Visual Tracking via Adaptive Structural Local Sparse Appearance Model},
	booktitle = {CVPR},
	year = {2012},
}

@inproceedings{ref:Kim,
	author = {J. Kim and D. Han and Y. W. Tai and J. Kim},
	title = {Salient Region Detection via High-Dimensional Color Transform},
	booktitle = {CVPR},
	year = {2014},
}

@inproceedings{ref:Boosting,
	author = {H. Grabner and M. Grabner and H. Bischof},
	title = {Real-Time Tracking via On-line Boosting},
	booktitle = {BMVC},
	year = {2006}
}

@inproceedings{ref:Wang,
	author = {L. Wang and H. Lu and X. Ruan and M.-H. Yang},
	title = {Deep Networks for Saliency Detection via Local Estimation and Global Search},
	booktitle = {CVPR},
	year = {2015},
}

@inproceedings{ref:Tong,
	author = {N. Tong and H. Lu and X. Ruan and M.-H. Yang},
	title = {Salient Object Detection via Bootstrap Learning},
	booktitle = {CVPR},
	year = {2015},
}

@inproceedings{ref:Li,
	author = {Changyang Li and Yuchen Yuan and Weidong Cai and Yong Xia and David Dagan Feng},
	title = {Robust Saliency Detection via Regularized Random Walks Ranking},
	booktitle = {CVPR},
	year = {2015},
}

@inproceedings{ref:Zhao,
	author = {R. Zhao and W. Ouyang and H. Li and X. Wang},
	title = {Saliency Detection by Multi-Context Deep Learning},
	booktitle = {CVPR},
	year = {2015},
}

@inproceedings{ref:Jiang2,
	author = {Huaizu Jiang and Jingdong Wang and Zejian Yuan and Yang Wu and Nanning Zheng and Shipeng Li},
	title = {Salient Object Detection: A Discriminative Regional Feature Integration Approach},
	booktitle = {CVPR},
	year = {2013},
}

@inproceedings{ref:Gong,
	author = {C. Gong and D. Tao and W. Liu and S. J. Maybank and M. Fang and K. Fu and J. Yang},
	title = {Saliency Propagation from Simple to Difficult},
	booktitle = {CVPR},
	year = {2015}
}

@inproceedings{ref:Li3,
	author = {X. Li and H. Lu and L. Zhang and X. Ruan and M.-H. Yang},
	title = {Saliency Detection via Dense and Sparse Reconstruction},
	booktitle = {ICCV},
	year = {2013},
}

@inproceedings{ref:Wang2,
	author = {Wenguan Wang and Jianbing Shen and F. Porikli},
	title = {Saliency-Aware Geodesic Video Object Segmentation},
	booktitle = {CVPR},
	year = {2015},
}

@article{ref:Hough,
	author = {J. Gall and A. Yao and N. Razavi and L. Van Gool and V. Lempitsky},
	title = {Hough Forests for Object Detection, Tracking, and Action Recognition},
	journal = {IEEE Trans. on PAMI},
	volume = {33},
	number = {11},
	year = {2011},
	pages = {2188-2202}
}

@inproceedings{ref:Grabner,
	author = {Helmut Grabner and Christian Leistner and Horst Bischof},
	title = {Semi-Supervised On-line Boosting for Robust Tracking},
	booktitle = {ECCV},
	year = {2008},
}

@inproceedings{ref:MKB,
	author = {F. Yang and H. Lu and Y. Chen},
	title = {Human Tracking by Multiple Kernel Boosting with Locality Affinity Constraints},
	booktitle = {ACCV},
	year = {2010},
}

@BOOK{ref:CART,
	author = {L. Breiman and J. Friedman and R. Olshen and C. Stone},
	title = {{Classification and Regression Trees}},
	publisher = {Wadsworth and Brooks},
	address = {Monterey, CA},
	year = {1984}
}

@inproceedings{ref:LongCT,
	author = {C. Ma and X. Yang and C. Zhang and M.-H. Yang},
	title = {Long-term Correlation Tracking},
	booktitle = {CVPR},
	year = {2015},
}

@inproceedings{ref:PGRF,
	author = {J. Choi and J. Y. Choi},
	title = {User Interactive Segmentation with Partially Growing Random Forest},
	booktitle = {ICIP},
	year = {2015},
}

@inproceedings{ref:Benchmark,
	author = {Y. Wu and J. Lim and M.-H. Yang},
	title = {Online Object Tracking: A Benchmark},
	booktitle = {CVPR},
	year = {2013},
}

@article{ref:Benchmark2,
	title = "Recent advances and trends in visual tracking: A review",
	journal = "Neurocomputing",
	volume = "74",
	number = "18",
	pages = "3823 - 3831",
	year = "2011",
	author = "Hanxuan Yang and Ling Shao and Feng Zheng and Liang Wang and Zhan Song"
}

@article{ref:Benchmark3,
	author = {Kevin Cannons},
	title = {A review of visual tracking},
	journal = {Dept. Comput. Sci. Eng., York Univ., Toronto, Canada, Tech. Rep. CSE-2008-07},
	year = {2008}
}

@article{ref:Benchmark4,
author={A. W. M. Smeulders and D. M. Chu and R. Cucchiara and S. Calderara and A. Dehghan and M. Shah},
journal={IEEE Trans. on PAMI},
title={Visual Tracking: An Experimental Survey},
year={2014},
volume={36},
number={7},
pages={1442-1468},
}


@article{ref:Center2,
	author = {Liu, Tie and Yuan, Zejian and Sun, Jian and Wang, Jingdong and Zheng, Nanning and Tang, Xiaoou and Shum, Heung-Yeung},
	title = {Learning to Detect a Salient Object},
	journal = {IEEE Trans. on PAMI},
	volume = {33},
	number = {2},
	year = {2011},
	pages = {353--367}
}

@inproceedings{ref:Chang,
	author = {H. J. Chang and H. Jeong and J. Y. Choi},
	title = {Active attentional sampling for speed-up of background subtraction},
	booktitle = {CVPR},
	year = {2012},
}

@article{ref:Lee_TIP2015,
	author = {K. Lee and D. Ognibene and H. J. Chang and T. K. Kim and Y. Demiris},
	title = {STARE: Spatio-Temporal Attention Relocation for Multiple Structured Activities Detection},
	journal = {IEEE Trans. on Image Processing},
	volume = {24},
	number = {12},
	year = {2015},
	pages = {5916-5927}
}

@inproceedings{ref:Sharma,
	author = {G. Sharma, F. Jurie, and C. Schmid},
	title = {Discriminative Spatial Saliency for Image Classification},
	booktitle = {CVPR},
	year = {2012}
}

@inproceedings{ref:Russakovsky,
	author = {O. Russakovsky, Y. Lin, K. Yu, and L. Fei-Fei},
	title = {Object-centric Spatial Pooling for Image Classification},
	booktitle = {ECCV},
	year = {2012},
}

@inproceedings{ref:Heo,
  author = {Heo, Byeongho and Jeong, Hawook and Kim, Jiyun and Choi, Sang-Il and Choi, Jin Young},
  booktitle = {ISVC (1)},
  pages = {647-657},
  series = {Lecture Notes in Computer Science},
  volume = 8887,
  year = 2014
}

@inproceedings{ref:Hong,
  title="Online Tracking by Learning Discriminative Saliency Map with Convolutional Neural Network",
  Author={Hong, Seunghoon and You, Tackgeun and Kwak, Suha and Han, Bohyung},
  year="2015",
  Booktitle="ICML",
  pages="597-606"
} 

@article{ref:Gauglitz,
	author="Gauglitz, Steffen and H{\"o}llerer, Tobias and Turk, Matthew",
	title="Evaluation of Interest Point Detectors and Feature Descriptors for Visual Tracking",
	journal="IJCV",
	year="2011",
	volume="94",
	number="3",
	pages="335"
}

@misc{ref:piotr,
   author = {Piotr Doll\'ar},
   title = {{P}iotr's {C}omputer {V}ision {M}atlab {T}oolbox ({PMT})},
   howpublished = {\url{https://github.com/pdollar/toolbox}}
} 

@article{ref:MIL,
	author = {B. Babenko and M.-H. Yang and S. Belongie},
	title = {Robust Object Tracking with Online Multiple Instance Learning},
	journal = {IEEE Trans. on PAMI},
	volume = {33},
	number = {8},
	year = {2011},
	pages = {1619-1632}
}

@inproceedings{ref:CT,
	author = {K. Zhang and L. Zhang and M.-H. Yang},
	title = {Real-time Compressive Tracking},
	booktitle = {ECCV},
	year = {2012},
}

@inproceedings{ref:CSK,
	author = {J. F. Henriques and R. Caseiro and P. Martins and J. Batista},
	title = {Exploiting the Circulant Structure of Tracking-by-detection with Kernels},
	booktitle = {ECCV},
	year = {2012},
}

@inproceedings{ref:DFT,
	author = {L. Sevilla-Lara and E. Learned-Miller},
	title = {Distribution Fields for Tracking},
	booktitle = {CVPR},
	year = {2012},
}

@article{ref:ALOV,
	author = {A. W. M. Smeulders and D. M. Chu and R. Cucchiara and S. Calderara and A. Dehghan and M. Shah},
	title = {Visual Tracking: An Experimental Survey},
	journal = {IEEE Trans. on PAMI},
	volume = {36},
	number = {7},
	year = {2014},
	pages = {1442-1468}
}

@article{ref:Felzenszwalb,
	author = {P. Felzenszwalb, R. Girshick, D. McAllester, and D. Ramanan},
	title = {Object detection with discriminatively trained part-based models},
	journal = {IEEE Trans. on PAMI},
	volume = {32},
	number = {9},
	year = {2010},
	pages = {1627-1645}
}

@inproceedings{ref:Lu,
	author = {Y. Lu and W. Zhang and C. Jin and X. Xue},
	title = {Learning Attention Map from Images},
	booktitle = {CVPR},
	year = {2012}
}

@article{ref:Kwon3,
	author = {J. Kwon and K. M. Lee},
	title = {Tracking by Sampling and Integrating Multiple Trackers},
	journal = {IEEE Trans. on PAMI},
	volume = {36},
	number = {7},
	year = {2014},
	pages = {1428-1441}
}

@article{ref:attention,
  author = {Yeshurun, Y. and Carrasco, M.},
  journal = {Nature},
  pages = 72,
  title = {Attention Improves or Impairs Visual Performance By Enhancing Spatial Resolution.},
  volume = 396,
  year = 1998
}

@inproceedings{ref:multi1,
	author = {Q. Bai and Z. Wu and S. Sclaroff and M. Betke and C. Monnier},
	title = {Randomized Ensemble Tracking},
	booktitle = {ICCV},
	year = {2013},
}

@inproceedings{ref:multi2,
	author = {J. Gao and H. Ling and W. Hu and J. Xing},
	title = {Transfer Learning Based Visual Tracking with Gaussian Processes Regression},
	booktitle = {ECCV},
	year = {2014},
}

@inproceedings{ref:multi3,
	author = {D. Lee and J. Sim and C. Kim},
	title = {Multihypothesis Trajectory Analysis for Robust Visual Tracking},
	booktitle = {CVPR},
	year = {2015},
}

@inproceedings{ref:AVSS2015,
	author = {B. Lee and K. Yun and J. Choi and J. Y. Choi},
	title = {Robust Pan-Tilt-Zoom Tracking via Optimization Combining Motion Features and Appearance Correlations},
	booktitle = {Advanced Video and Signal Based Surveillance (AVSS)},
	year = {2015},
}

@inproceedings{ref:AVSS2015,
	author = {B. Lee and K. Yun and J. Choi and J. Y. Choi},
	title = {Robust Pan-Tilt-Zoom Tracking via Optimization Combining Motion Features and Appearance Correlations},
	booktitle = {Advanced Video and Signal Based Surveillance (AVSS)},
	year = {2015},
}

@inproceedings{ref:SCT,
  title={Visual Tracking Using Attention-Modulated Disintegration and Integration},
  author={Jongwon Choi and Hyung Jin Chang and Jiyeoup Jeong and Yiannis Demiris and Jin Young Choi},
  booktitle = {CVPR},
	year = {2016},
}

@inproceedings{ref:COT,
  title={Beyond Correlation Filters: Learning Continuous Convolution Operators for Visual Tracking},
  author={Martin Danelljan and Andreas Robinson and Fahad Shahbaz Khan and Michael Felsberg},
  booktitle = {ECCV},
	year = {2016},
}

@inproceedings{ref:DeepSRDCF,
  title={Convolutional Features for Correlation Filter Based Visual Tracking},
  author={Martin Danelljan and Gustav Hager and Fahad Shahbaz Khan and Michael Felsberg},
  booktitle = {ICCV workshops},
	year = {2016},
}

@inproceedings{ref:FCNT,
  title={Visual Tracking with Fully Convolutional Networks},
  author={Lijun Wang and Wanli Ouyang and Xiaogang Wang and Huchuan Lu},
  booktitle = {ICCV},
	year = {2015},
}

@inproceedings{ref:MDNet,
  title={Learning Multi-Domain Convolutional Neural Networks for Visual Tracking},
  author={Hyeonseob Nam and Bohyung Han},
  booktitle = {CVPR},
	year = {2016},
}


@article{ref:HAMMER,
  author = {Yiannis Demiris and Bassam Khadhouri},
  journal = {Robotics and Autonomous Systems},
  pages = {361-369},
  title = {Hierarchical Attentive Multiple Models for Execution and Recognition of Actions},
  volume = 54,
  number = 5,
  year = 2006
}

@article{ref:LSTM,
  author = {Sepp Hochreiter and Jurgen Schmidhuber},
  journal = {Neural Computation},
  pages = {1735-1780},
  title = {Long Short-term Memory},
  volume = 9,
  number = 8,
  year = 1997
}

@article{ref:nature1,
  author = {Andreas K. Engel and Pascal Fries and Wolf Singer},
  journal = {Nature Reviews Neuroscience},
  pages = {704-716},
  title = {Dynamic Predictions: Oscillations and Synchrony in Top–down Processing},
  volume = 2,
  year = 2001
}

@inproceedings{ref:adam,
  title={{Adam: A Method for Stochastic Optimization}},
  author={Diederik Kingma and Jimmy Ba},
  booktitle = {International Conference for Learning Representations},
	year = {2015},
    pages = {}
}

@article{gilbert2013top,
  title={Top-down influences on visual processing},
  author={Gilbert, Charles D and Li, Wu},
  journal={Nature Reviews Neuroscience},
  volume={14},
  number={5},
  pages={350--363},
  year={2013},
}

@book{tsotsos2011,
  title={A computational perspective on visual attention},
  author={Tsotsos, John K},
  year={2011},
  publisher={MIT Press}
}

@InProceedings{ref:vot2014,
author="Kristan, Matej
and others",
title="The Visual Object Tracking VOT2014 Challenge Results",
bookTitle="ECCV 2014 Workshop",
year="2015",
}

@InProceedings{ref:vot2015,
author = {Kristan, Matej and others},
title = {{The Visual Object Tracking VOT2015 Challenge Results}},
booktitle = {{ICCV Workshops}},
year = {2015},
pages={1--23}
} 

@InProceedings{ref:vot2016,
author = {Kristan, Matej and others},
title = {{The Visual Object Tracking VOT2016 Challenge Results}},
booktitle = {{ICCV Workshops}},
year = {2016}
} 

@article{ref:TPAMI2015Benchmark,
author={Y. Wu and J. Lim and M.-H. Yang},
journal={IEEE Trans. on PAMI},
title={Object Tracking Benchmark},
year={2015},
volume={37},
number={9},
pages={1834-1848},
}

@inproceedings{ref:SINT,
  title={Siamese Instance Search for Tracking},
  author={Ran Tao and Efstratios Gavves and Arnold W.M. Smeulders},
  booktitle = {CVPR},
	year = {2016},
}

@inproceedings{ref:STCT,
  title={{STCT}: Sequentially Training Convolutional Networks for Visual Tracking},
  author={Lijun Wang and Wanli Ouyang and Xiaogang Wang and Huchuan Lu},
  booktitle = {CVPR},
	year = {2016},
}

@inproceedings{ref:CF2,
  title={Hierarchical Convolutional Features for Visual Tracking},
  author={Chao Ma and Jia-Bin Huang and Xiaokang Yang and Ming-Hsuan Yang},
  booktitle = {ICCV},
	year = {2015},
}

@inproceedings{ref:HDT,
  title={Hedged Deep Tracking},
  author={Yuankai Qi and Shengping Zhang and Lei Qin and Hongxun Yao and Qingming Huang and Jongwoo Lim and Ming-Hsuan Yang},
  booktitle = {CVPR},
	year = {2016},
}

@article{ref:haefner,
  title={Perceptual Decision-Making as Probabilistic Inference by Neural Sampling},
  author={Ralf M. Haefner and Pietro Berkes and József Fiser},
  journal={Neuron},
  volume={90},
  number={3},
  pages={649--660},
  year={2016},
}

@article{ref:hiroshi,
  title={Learning enhances the relative impact of top-down processing in the visual cortex},
  author={Hiroshi Makino and Takaki Komiyama},
  journal={Nature Neuroscience},
  volume={18},
  pages={1116--1122},
  year={2015},
}

@inproceedings{ref:SRDCF,
  title={Learning Spatially Regularized Correlation Filters for Visual Tracking},
  author={Martin Danelljan and Gustav Häger and Fahad Khan and Michael Felsberg},
  booktitle = {ICCV},
	year = {2015},
}

@inproceedings{ref:SiamFC,
  title={Fully-Convolutional Siamese Networks for Object Tracking},
  author={Luca Bertinetto and Jack Valmadre and João F. Henriques and Andrea Vedaldi and Philip H.S. Torr},
  booktitle = {ECCV workshop},
	year = {2016},
}

@article{ref:ImageNet,
  title={ImageNet Large Scale Visual Recognition Challenge},
  author={Olga Russakovsky and Jia Deng and Hao Su and Jonathan Krause and Sanjeev Satheesh and Sean Ma and Zhiheng Huang and Andrej Karpathy and Aditya Khosla and Michael Bernstein and Alexander C. Berg and Li Fei-Fei},
  journal={IJCV},
  volume={115},
  number={3},
  pages={211--252},
  year={2015},
}

@article{ref:CullyNature,
  title={Robots that can adapt like animals},
  author={Cully, Antoine and Clune, Jeff and Tarapore, Danesh and Mouret, Jean-Baptiste},
  journal={Nature},
  volume={521},
  number={7553},
  pages={503--507},
  year={2015}
}

@article{ref:Stare,
  title={STARE: Spatio-Temporal Attention RElocation for Multiple Structured Activities Detection},
  author={Kyuhwa Lee and Dimitri Ognibene and Hyung Jin Chang and Tae-Kyun Kim and Yiannis Demiris},
  journal={Transactions on Image Processing},
  volume={24},
  number={12},
  pages={5916--5927},
  year={2015},
}

@article{Demiris2014,
author = {Demiris, Yiannis and Aziz-Zadeh, Lisa and Bonaiuto, James},
number = {1},
pages = {63--91},
title = {{Information Processing in the Mirror Neuron System in Primates and Machines}},
volume = {12},
journal = {Neuroinformatics},
year = {2014}
}

@incollection{Demiris2002,
author = {Demiris, Yiannis},
pages = {327--361},
booktitle={Imitation in Animals and Artifacts},
title={Imitation as a dual-route process featuring predictive and learning components: a biologically-plausible computational model},
year={2002},
publisher = {MIT Press},
}

@article{Frintrop2010,
 author = {Frintrop, Simone and Rome, Erich and Christensen, Henrik I.},
 title = {Computational Visual Attention Systems and Their Cognitive Foundations: A Survey},
 journal = {ACM Trans. Appl. Percept.},
 volume = {7},
 number = {1},
 year = {2010},
 pages = {6:1--6:39},
} 

@article{ref:sparsity,
 author = {Weston, Jason and Elisseeff, Andr{\'e} and Sch\"{o}lkopf, Bernhard and Tipping, Mike},
 title = {Use of the Zero Norm with Linear Models and Kernel Methods},
 journal = {J. Mach. Learn. Res.},
 issue_date = {3/1/2003},
 volume = {3},
 year = {2003},
 issn = {1532-4435},
 pages = {1439--1461},
 numpages = {23},
 url = {http://dl.acm.org/citation.cfm?id=944919.944982},
 acmid = {944982},
 publisher = {JMLR.org},
} 

@incollection{ref:lecun,
  title={Efficient backprop},
  author={LeCun, Yann A and Bottou, L{\'e}on and Orr, Genevieve B and M{\"u}ller, Klaus-Robert},
  booktitle={Neural networks: Tricks of the trade},
  pages={9--48},
  year={2012},
}

@misc{ref:tensorflow,
title={{TensorFlow}: Large-Scale Machine Learning on Heterogeneous Systems},
url={http://tensorflow.org/},
note={Software available from tensorflow.org},
author={
    Mart\'{\i}n~Abadi and others},
  year={2015},
}
}

\end{document}